# Swarm Intelligence for Self-Organized Clustering

**Michael C. Thrun**                                    **Alfred Ultsch**

*mthrun@mathematik.uni-marburg.de*          *ultsch@Mathematik.Uni-Marburg.de*

*Databionics Research Group,*
*Philipps-University of Marburg,*
*Hans-Meerwein-Straße 6,*
*D-35032 Marburg, Germany*

**Editor:**

## Abstract

Algorithms implementing populations of agents which interact with one another and sense their environment may exhibit emergent behavior such as self-organization and swarm intelligence. Here a swarm system, called Databionic swarm (DBS), is introduced which is able to adapt itself to structures of high-dimensional data characterized by distance and/or density-based structures in the data space. By exploiting the interrelations of swarm intelligence, self-organization and emergence, DBS serves as an alternative approach to the optimization of a global objective function in the task of clustering. The swarm omits the usage of a global objective function and is parameter-free because it searches for the Nash equilibrium during its annealing process. To our knowledge, DBS is the first swarm combining these approaches. Its clustering can outperform common clustering methods such as K-means, PAM, single linkage, spectral clustering, model-based clustering, and Ward, if no prior knowledge about the data is available. A central problem in clustering is the correct estimation of the number of clusters. This is addressed by a DBS visualization called topographic map which allows assessing the number of clusters. It is known that all clustering algorithms construct clusters, irrespective of the data set contains clusters or not. In contrast to most other clustering algorithms, the topographic map identifies, that clustering of the data is meaningless if the data contains no (natural) clusters. The performance of DBS is demonstrated on a set of benchmark data, which are constructed to pose difficult clustering problems and in two real-world applications.

# 1    Introduction

No generally accepted definition of clusters exists in the literature [C. Hennig, et al. (Hg.), 2015, p. 705]. Additionally, Kleinberg showed for a set of three simple properties (scale-invariance, consistency and richness), that there is no clustering function satisfying all three [Kleinberg, 2003] where the property of richness implies, that all partitions should be achievable. Therefore, the focus lies here on the concept of natural clusters (cf. [Duda et al., 2001, p. 539; Theodoridis/Koutroumbas, 2009, pp. 579, 600]). Natural clusters are characterized by distances and/or density-based structures [Duda et al., 2001; Theodoridis/Koutroumbas, 2009]. Then, the purpose of clustering methods is to identify homogeneous groups of objects (cf. [Arabie et al., 1996, p. 8]), such that objects are similar within clusters and dissimilar between clusters. However, many clustering algorithms implicitly assume different structures of clusters [Duda et al., 2001, pp. 537, 542, 551; Everitt et al., 2001, pp. 61, 177; Handl et al., 2005; Theodoridis/Koutroumbas, 2009, pp. 896, 896; Ultsch/Lötsch, 2017]:

"[Natural] [c]lusters can be of arbitrary shapes (structures) and sizes in a multidimensional pattern space. Each clustering criterion imposes a certain structure on the data, and if the data happen to conform to the requirements of a particular criterion, the true clusters are recovered. Only a small number of independent clustering criteria can be understood both mathematically and intuitively. Thus the hundreds of criterion functions proposed in the literature are related and the same criterion appears in several disguises" [Jain/Dubes, 1988, p. 91].

On a benchmark of several artificial datasets, this work will demonstrate that conventional clustering algorithms are limited in their clustering ability in the presence clusters defined by a combination of distance- and density-based structures. The conclusion is that when the task is to achieve a structure-preserving clustering, the optimization of an objective function could yield misleading results if the underlying structures of the high-dimensional data of interest are unknown.

Hence, a completely different approach is required, which motivates an extensive review of the application of artificial intelligence in data science in the next section. There, two interesting concepts are addressed, called self-organization and swarm intelligence. Through self-organization, the irreducible structures of high-dimensional data can emerge, in a process that can lead to emergence. If properly applied using a swarm of intelligent agents, the approach presented in this work can outperform the optimization of an objective function for the task of clustering. Section two will outline the reasons why the Databionic swarm will only be compared to conventional clustering algorithms. First, swarm algorithms are mainly hybrids with common clustering methods and therefore it is assumed that they will seek and find the same structure types. Second, there is currently no open-source code available except for applications of rule-based classification [Martens et al., 2011]. Third, the authors argue that comparison with state of the art approaches is sufficient for the structure argument taken above. The Databionic Swarm (DBS) proposed in this work consists of three independent modules. The result of the first module is the parameter-free projection of high-dimensional data into a two-dimensional space without the need for a global objective function. However, the Johnson–Lindenstrauss lemma [Dasgupta/Gupta, 2003] states, that the two-dimensional similarities in the scatter plot cannot coercively represent high-dimensional distances. The second module accounts for the errors in the projection for which the Generalized U-Matrix technique is applied [Ultsch/Thrun, 2017]. It generates a topographic map with hypsometric tints [Thrun et al., 2016]. The visualized data structures can even be clustered manually by the user using the interactive interface of shiny (https://shiny.rstudio.com/). In the third module, the automatic clustering approach is proposed in this work because it enables to compare many trials of DBS to conventional clustering algorithms. A detailed description of DBS can be found in section three.

## 2    Behavior based systems in unsupervised machine learning

Many technological advances have been achieved with the help of bionics, which is defined as the application of biological methods and systems found in nature. A related, rarely discussed subfield of information technology is called databionics. Databionics refers to the attempt to adopt information processing techniques from nature. This section will discuss the imitation of natural processes (also called biomimicry [Benyus, 2002]) using swarm intelligence, which is a form of artificial intelligence (AI) [Bonabeau et al., 1999] and was introduced as a term in the context of robotics [Beni/Wang, 1989].

Swarm intelligence is defined as the emergent collective behavior of simple entities called agents [Bonabeau et al., 1999, p. 12]. In the context of swarms, the terms behavior and intelligence are used synonymously, bearing in mind that in general, the definition of intelligence is controversial [Legg/Hutter, 2007] and complex [Zhong, 2010].

Swarm behavior can be imitated based on observations of herds [Wong et al., 2014], bird flocks and fish schools [Reynolds, 1987], bats [Yang/He, 2013], or insects such as bees [Karaboga, 2005; Karaboga/Akay, 2009], ants [Deneubourg et al., 1991], fireflies [Yang, 2009], cockroaches [Havens et al., 2008], midges [Passino, 2013], glow-worms or slime moulds [Parpinelli/Lopes, 2011]. [Grosan et al.] define five main principles of swarm behavior: Homogeneity, meaning that every agent has the same behavior model; Locality, meaning that the motion of each agent is influenced only by its nearest neighbors; Velocity Matching, meaning that each agent attempts to match the velocity of nearby flockmates; Collision Avoidance, meaning that each agent avoids collisions with nearby agents; and Flock Centering, meaning that agents attempt to stay close to neighboring agents [Reynolds, 1987, pp. 6, 7; Grosan et al., 2006, p. 2]. Here, these definitions are given greater specificity in two respects. First, the term agent is modified to the term agents of the same type because many swarms consists of more than one type of agent, e.g., small and large workers in the Pheidole genus of ants [Bonabeau et al., 1999, p. 4]. Second, a swarm need not necessarily to move. For example, fire ants self-assemble into waterproof rafts to survive floods [Mlot et al., 2011]. The individual ants are linked together to construct such self-assemblages [Mlot et al., 2011]. Therefore, velocity matching can result in a velocity of zero.

If a swarm contains a sufficient number of agents, self-organization may emerge. Self-organization is defined as the spontaneous formation of patterns by a system itself [Kelso, 1997, p. 8 ff.], without any central control. During self-organization, novel and coherent structures, patterns, and properties may arise [Goldstein, 1999]. This ability of a system to produce phenomena on a new, higher level is called emergence [Ultsch, 1999], and it is separately discussed in section 2.1.

"Self-organizing swarm behavior relies on four basic ingredients" [Bonabeau et al., 1999, pp. 22-25]: positive feedback, negative feedback, amplification of fluctuations and multiple interactions. The first two factors promote the creation of convenient structures and help to stabilize them. Fluctuations are defined to include errors, random movements and task switching. For swarm behavior to emerge, multiple interactions between agents are required. Agents can communicate with each other either directly or indirectly. An example of direct communication is the dancing behavior of bees, in which a bee shares information about a food source, such as how plentiful it is and its direction and distance away [Karaboga/Akay, 2009]. Indirect communication is observed, for example, in the behavior of ants [Schneirla, 1971]. If the agents communicate only through modifications to their environment (through pheromones, for example), then this type of communication is defined as stigmergy [Grassé, 1959; Beckers et al., 1994].

The exact number of agents required for self-organization is unknown, but it should be not so large that it must be handled in terms of statistical averages and not so small that it can be treated as a few-body problem [Beni, 2004]. For example, 4096 neurons are required for self-organization in SOMs [Ultsch, 1999], and for the coordinated marching behavior of locusts, a minimum density of least 30 agents (73.8 locusts/m^2 in vivo experiments) was reported in [Buhl et al., 2006, p. 1404] which can be seen as self-organization in-between agents. Considering the two requirements stated above, Beni defined a swarm as a formation of cellular robots with a number exceeding 100 [Beni, 2004]. Here, consistent with [Beni, 2004], the argument is made that for self-organization in a swarm, the number of agents should be higher than 100.

The two main types of swarm-based analysis discussed in data science; namely, particle swarm optimization (PSO) and ant colony systems (ACS) [Martens et al.], are distinguished by the type of communication used: PSO agents communicate directly, whereas ACS agents communicate through stigmergy. PSO methods are based on the movement strategies of particles [Kennedy/Eberhart, 1995] and typically used as population-based search algorithms [Rana et al., 2011], whereas ACS methods are applied for sorting tasks [Martens et al., 2011].

In addition to being used to solve discrete optimization problems, PSO has been used as a basis for rule-based classification models [Wang et al., 2007], or as an optimizer within other learning algorithms [Martens et al., 2011], whereas ACS has been used primarily for classification within the data mining community [Martens et al., 2011]. Pseudocode for both types of algorithms and illustrative descriptions can be found in [Abraham et al., 2006].

For clustering tasks, PSO has mainly been applied in hybrid algorithms [Esmin et al., 2015]; e.g., [Van der Merwe/Engelbrecht, 2003] applied PSO combined with k-means clustering. Here, it is argued that the hybridization of PSO and k-means may improve the choice of centroids or may, in some special cases, even allow the problem of the number of clusters to be solved. However, this approach is subject to several of the shortcomings of k-means, which is known to search for spherical clusters [C. Hennig, et al. (Hg.), 2015, p. 721], i.e., it is unable to find clusters in elementary data sets, such as those in the Fundamental Clustering Problems Suite (FCPS) [Thrun/Ultsch, 2020a] (see also Fig. 5.1).

According to [Rana et al., 2011], the advantages of the clustering process are in the case of the PSO approach that it is very fast, simple and easy to understand and implement. "PSO also has very few parameters to adjust [Eberhart et al., 2001] and requires little memory for computation. Unlike other evolutionary and mathematical algorithms it is more computationally effective" [Rana et al., 2011] (citing [Arumugam et al., 2005]). Again according to [Rana et al., 2011], the disadvantages are the "poor quality results when it deals with large and complex data sets". "PSO gives good results and accuracy for single-objective optimization, but for a multi-objective problem, it becomes stuck in local optima" [Rana et al., 2011] (citing [Li/Xiao, 2008]). Another problem with PSO is its tendency to reach fast and premature convergence at mid-optimum points [Rana et al., 2011]. It is difficult to find the correct stopping criterion for PSO [Bogon, 2013], which is usually one of the following: a fixed maximum number of iterations, a maximum number of iterations without improvement or a minimum objective function error [Abraham et al., 2006; Esmin et al., 2015]. Hybrid PSO algorithms usually optimize an objective function [Bogon, 2013, pp. 39ff, 46] and therefore always make implicit assumptions regarding the underlying structures of the data. Notably, there is no single "best" criterion for obtaining a clustering because no precise and workable definition of "a cluster" exists [Jain/Dubes, 1988, p. 91].

ACS methods for clustering tasks are divided into ant colony optimization (ACO) and ant-based clustering (ABC) (for an overview, see [Kaur/Rohil, 2015]). "Like ant colony optimisation (ACO), ant-based clustering and sorting is a distributed process that employs positive feedback. However, in contrast to ACO, no artificial pheromones are used; instead, the environment itself serves as stigmergic variable" [Handl et al., 2003]. ACO has been directly applied to the task of clustering in [Shelokar et al., 2004; Menéndez et al., 2014] using global objective functions similar to k-means. ABC methods model the behavior of ant colonies, and data points are picked up and dropped off accordingly [Bonabeau et al., 1999]. ABC was introduced by [Deneubourg et al., 1991] as a way to explain the phenomenon of the gathering and sorting of corpses observed among ants. In an experiment the ants formed cemeteries of dead ants that had been randomly scattered beforehand.

[Deneubourg et al., 1991] proposed probability functions for the picking up and dropping off of the corpses. Because ants are very specialized in their roles, several different types of ants of the same species exist in a colony, and different individuals in the colony perform different tasks. The probabilities are calculated as functions of the number of corpses of the same type in a nearby area (positive feedback). For a clustering task, the ants and data points (representing ant corpses) are randomly placed on a grid, and the ants move randomly across the grid, at times picking up and carrying the data points [Lumer/Faieta, 1994]. The probabilities of picking up and dropping off the data points are modified according to a dissimilarity-based evaluation of the local density (see [Kaur/Rohil, 2015] and [Jafar/Sivakumar, 2010], citing [Lumer/Faieta, 1994]).

[Handl et al., 2006] enhanced the algorithm; they called their version Adaptive Time-dependent Transporter Ants (ATTA) because they incorporated adaptive heterogeneous ants and time-dependent transport activities into the algorithm. Further improvements to the picking up and dropping off activities were presented in [Ouadfel/Batouche, 2007; Omar et al., 2013], and improvements to the initialization and post-processing were proposed in [Aparna/Nair, 2014]. Another version of the approach was developed by introducing an annealing scheme [Tsai et al., 2004].

The main problem in ABC lies in the fact that the picking up and dropping off behaviors are independent of the number of agents required to execute the task [Tan et al., 2006; Herrmann/Ultsch, 2008a, 2008b, 2009; Herrmann, 2011, p. 81]. Furthermore, Tan et al showed that collective intelligence of ABC can be discarded and the same results can be achieved using an objective function without ants [Tan et al., 2006; Herrmann/Ultsch, 2008b] which means ABC methods can be regarded as derived from the batch-SOM algorithm [Herrmann/Ultsch, 2008a]. From this perspective, an ABC algorithm possesses an objective function, which can be decomposed into an output density term multiplied by one minus a topographic quality term called stress [Herrmann/Ultsch, 2008a, p. 3; 2008b, p. 217; 2009, p. 4; Herrmann, 2011, pp. 137-138]. Both terms are minimized simultaneously [Herrmann/Ultsch, 2008a, 2008b, 2009]. However, the output density term is easy to optimize but distorts the correct clustering of the data. Consequently, the above-mentioned authors used this mathematical stress term to define a scent as follows:

Let D(l, j) be the distance between two points $x_l, x_j \in I$, let d(l, j) be the corresponding distance in the output space O, and let $h_R: R \rightarrow [0,1]$ be an arbitrary but continuous and monotonically decreasing function; then, the scent $\lambda(b_j, R): \mathbb{R}_0^+ \times O \rightarrow \mathbb{R}_0^+$ is defined as

$$\lambda(b_j, R) = \frac{\sum_{l \in I} h_R(d(j,l)) * D(j,l)}{\sum_{l \in I} h_R(d(j,l))} \qquad (1)$$

The scent $\lambda$ is the weighted sum of the distances to neighboring objects using an unspecified neighborhood function $h_R$.

In sum, Beni argued that at least 100 agents are required to make a swarm [Beni, 2004] and, to our knowledge, self-organization was not observed below 30 agents [Buhl et al., 2006, p. 1404]. However, only one agent is required in ABC methods, and consequently, swarm-based self-organization of ABC- algorithms is controversial.

In unsupervised learning, two additional approaches are known. Methods of the first type are founded on an analysis of the behavior of bees [Karaboga, 2005]. These are hybrid approaches to clustering that use swarm intelligence in combination with other methods, e.g., k-means [Marinakis et al., 2007; Pham et al., 2007; Zou et al., 2010; Karaboga/Ozturk, 2011] or SOM [Fathian/Amiri, 2008].

Methods of the second type are based on foraging theory, which focuses on two basic problems: which prey a forager should consume and when a forager should leave a patch [Stephens/Krebs, 1986, p. 6]. A forager is viewed as an agent who compares a potential energy gain with a potential opportunity for finding an item of a superior type [Martens et al., 2011] (citing [Stephens/Krebs, 1986]). This approach is also called the prey model [Martens et al., 2011]: the average energy gain can be mathematically expressed in terms of the expected time, energy intake, encounter rate and attack probability for each type of prey. In the projection method proposed by [Giraldo et al., 2011], in addition to the characteristics of the approach described above, the "foraging landscape was viewed as a discrete space, and objects representing points from the dataset as prey." There were three agents defined as foragers.

## 2.1    Missing Links: Emergence and Game Theory

Through self-organization, novel and irreducible[1] structures, patterns, and properties can emerge in a complex system [Goldstein, 1999]. In analogy to SOMs [Ultsch, 1999], this idiosyncratic behavior of a swarm is defined here as *emergence*.

Sometimes, a distinction is made between strong and weak emergence [Janich/Duncker, 2011, p. 19]. Here, only strong emergence is relevant. In the literature, the existence of emergence is controversial[2]; it is possible that the concept is only required because the causal explanations for certain phenomena have not yet been found [Janich/Duncker, 2011, p. 23]. Figure 2.1 presents an example of emergence in swarms. The non-deterministic movement of fish is temporarily and structurally unpredictable and consists of many interactions among many agents. Nevertheless, this fish school forms a ball-like formation. It appears that the concept of emergence has remained unused and rarely discussed in the literature on swarm intelligence, although it is a key concept in AI [Brooks, 1991]. Emergence is mentioned in the literature as a biological aspect of swarms [Garnier et al., 2007], in distributed AI for complex optimization problems [Bogon, 2013, p. 19],in the context of software systems [Bogon, 2013, p. 19] (citing [Timm, 2006]) and as emergent computation [Forest, 1990].

Contrary to Forest, who assumes that only cooperative behavior can lead to emergence [Forest, 1990, p. 8], this works shows that egoistic behavior of a swarm can lead to emergence as well.

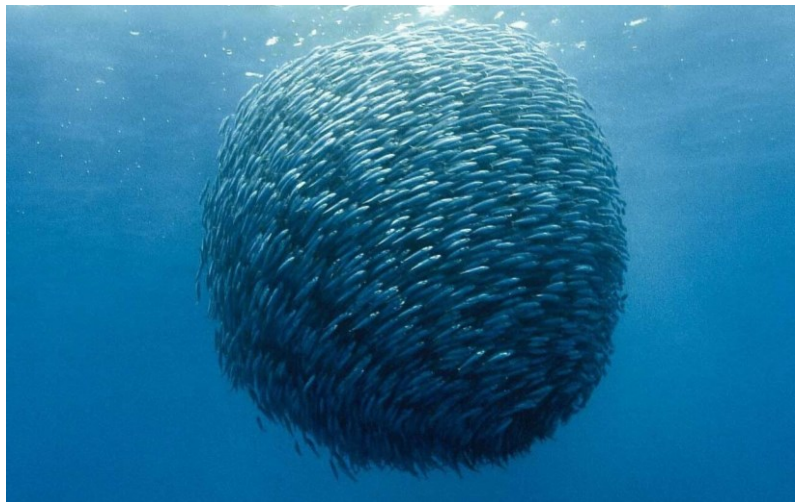

Figure 2.1: A fish swarm in the form of a ball [Uber_Pix, 2015]: an example of emergence in swarms.

---

[1] There is no way to derive the property from any part, subset or partial structure of the system [Ultsch, 2007].

[2] For applications, the existence of emergence is irrelevant. Even if emergent phenomena can be causally explained, they can still be used in the future.

With regard to swarms, emergence should be a key concept. The four factors leading to emergence in swarms are

  I.    Randomness
  II.   Temporal and structural unpredictability
  III.  Multiple **non-linear** interactions among **many** agents
  IV.   Irreducibility

[Bonabeau et al., 1999, p. 23] agrees with [Ultsch, 1999, 2007] regarding the first factor: "*Randomness* is often crucial, since it enables the discovery of new solutions, and *fluctuations* can act as seeds from which structures nucleate and grow." Here, an algorithm is considered to have the property of *randomness* if it uses a source of random numbers in its calculations (non-determinism) [Ultsch, 2007]. The power of randomness is evident in Schelling's segregation model[Schelling, 1969].

The second factor, unpredictability [Ultsch, 2007], is incompatible with the ACO and PSO approach, in which an objective function is optimized [Martens et al., 2011] and, therefore, predictable assumptions are implicitly made regarding the structures of data sets in the case of unsupervised machine learning (see [Thrun, 2018] for further details).

The third factor, multiple interactions among many agents, was identified by [Forest, 1990, pp. 1-2] for nonlinear systems. Although [Bonabeau et al., 1999] defines a requirement of multiple interactions for self-organization, the authors argue on page 24 that a single agent may also be sufficient. This is not the case for emergence, for which many elementary processes are mandatory [Ultsch, 1999; Beni, 2004]. Hence, ABC methods cannot exhibit the property of emergence. Nonlinearity means that adding or removing interactions among agents or any agents themselves results in behavior that is linearly unpredictable. For example, the removal of one DataBot results in the elimination of one data point.

The fourth factor, Irreducibility [Kim, 2006, p. 555, Ultsch, 2007, O'Connor/Wong, 2015], means that the (novel) property cannot be derived from any agent (or part) of the system, but is only a property of the whole system. It is the ability of a system to produce phenomena on a new, higher level [Ultsch, 1999].

The second missing link is a connection to game theory, in which the four axioms of self organization — *positive* and *negative feedback*, *amplification of fluctuations* and *multiple interactions* — are apparent. Game theory was introduced by [Neumann/Morgenstern] in 1947. The purpose of game theory is to model situations[3] in which multiple players interact with each other or affect each other's outcomes [Nisan et al., 2007, p. 3] (*multiple interactions*). Here, the focus lies on a general, not zero-sum, and n-person game [Neumann/Morgenstern, 1953, p. 85].

---

[3] To be more specific, rational decision-making behavior in social conflict situations.

A game is defined as a scenario with n players i=1, …, n in which each player makes a choice [Neumann/Morgenstern, 1953, p. 84] (*amplification of fluctuations*[4]).

Let a game $G$ be defined by $n$ players associated with $n$ non-empty sets $\Pi_1, \dots, \Pi_n$, where every set $\Pi_i$ represent all choices made by player $i$; then, the pay-off function is defined as

$$p = (p_1, \dots, p_n): \Pi_1 \times \dots \times \Pi_n \to \mathbb{R}^n$$

The choices of each player determine the outcome for each player, and the outcome will, in general, be different for different players [Nisan et al., 2007, p. 9]. In a game, the payoff for each player depends on not only his own choices but also the choices of all other players [Nisan et al., 2007, p. 9] (*positive* and *negative feedback*). Often, the choices are defined based on a set of mixed strategies for each player. From the biological point of view, these mixed strategies may include the five main principles of collective behavior: *Homogeneity*, *Locality*, *Velocity Matching*, *Collision Avoidance*, and *Flock Centering* [Grosan et al., 2006].

In a game with $n$ players, let the k choices of player $i$ be defined by a set $\Pi_i = \{\pi_1^i, \dots \pi_\alpha^i, \dots, \pi_k^i\}$, where $\pi_\alpha^i$ indicates the $i^{th}$ player's $\alpha^{th}$ choice; then, a mixed strategy $s_j(i) \in S_i$ for the player $i$ is defined by

$$s_j(i) = \sum_{\alpha=1}^{k(i)} c_\alpha(i)\, \pi_\alpha(i)$$

where

$$\sum_{\alpha=1}^{k(i)} c_\alpha(i) = 1 \text{ and all } c_\alpha(i) \geq 0.$$

For noncooperative games, [Nash, 1951] proved the existence of at least one equilibrium point. Let $t_j(i) \in S_i$ be the mixed strategy that maximizes the payoff for the player $i$; then, the Nash equilibrium is defined as

$$p_i\big(s(1),\dots,s(i-1),t_j(i),s(i+1),\dots,s(n)\big) = \max_{t_j(i) \in S_i} p_i\big(s(1),\dots,s(n)\big) \qquad (2)$$

if and only if this equation holds for every i [Nash, 1951]. The mixed strategy $t_j(i) \in S_i$ is the equilibrium point if no deviation in strategy by any single person results in a greater profit for that person. A Nash equilibrium is called weak if multiple mixed strategies $t_j(i) \in S_i$ for the same person exist in equation (2) that result in the same maximal payoff $p_i$, whereas in a strong Nash equilibrium, even a coalition of players cannot further increase their payoffs by simultaneously changing their strategies $t_j(i) \in S_i, i = 1 \dots m \leq n$, in (2).

---

[4] Task switching.

### 3    Databionic Swarm (DBS)

An agent is a software entity, situated in a given environment, that is capable of flexible, autonomous action in order to meet its design objectives [Jennings et al., 1998]. In the context of data science, the first agents to be developed and applied were called DataBots [Ultsch, 2000a]. Each DataBot possesses a scent, defined by one high-dimensional data point and DataBots possess probabilistically defined movement strategies:

Let each DataBot $b_j \in B$ be an agent identified by a numerical vector $z_j \epsilon \mathbb{R}^d$; it resides on a large, finite, two-dimensional discrete grid that can be embedded on the surface of a torus [Ultsch, 2000a]. The current position of DataBot $b_j$ is denoted by $w_j \epsilon$ O. Every DataBot $b_j = \{w_j, z_j\}$ emits a scent $\lambda$, which is detected by all other DataBots in its neighborhood.

The swarm of DataBots introduced here, is called a polar swarm (Pswarm) because its agents move in polar coordinates based on symmetry considerations (see [Feynman et al., 2007, p. 745]). All parameters are automatically chosen according to, and directly based on, the appropriate high-dimensional definition of distance. The main idea of Pswarm is to combine the concepts of swarm intelligence and self-organization with non-cooperative game theory [Nash, 1950]. The main advance is the reliance on the concept of emergence [Ultsch, 2007] instead of the optimization of an objective function. This allows Pswarm to preserve structures in data sets that are characterized by distance and/or density.

The extensive analysis of ant-based clustering (ABC) methods that have been performed in previous work allows the formulation of a precise mathematical definition of pheromonal stigmergy (scent) [Herrmann/Ultsch, 2009]. The scent is defined in each neighborhood using an annealing scheme. The approach based on neighborhood reduction during the annealing process was invented by Kohonen [Kohonen, 1982] and was used, for example, in [Demartines/Hérault, 1995; Ultsch, 1999; Hinton/Roweis, 2002]. In the context of swarm-based techniques, it was used for the first time in [Tsai et al., 2004]. Until now, finding the correct annealing scheme for a high-dimensional data set has remained a challenging task [Nybo et al., 2007]. The Pswarm algorithm utilizes randomness and the Nash equilibrium [Nash, 1950] of non-cooperative game theory to find an appropriate annealing scheme based on the data as given in the input space. For this purpose, the scent will be redefined as the payoff function.

Having projected the high-dimensional points into two dimensions using Pswarm in section 3.1, the author applies the generalized U-matrix [Ultsch/Thrun, 2017], to propose a three-dimensional topographic map with hypsometric tints [Thrun et al., 2016] based on the high-dimensional distances and the density of the two-dimensional projected points. Drawing further insights from [Lötsch/Ultsch, 2014], a semi-interactive, but parameter-insensitive, clustering approach is possible. The framework as a whole is called Databionic swarm (DBS) and has only two parameters: the number of clusters and the type of clustering (connected or compact). The key feature of DBS is that neither an overall objective function for the process nor the type of clusters sought is explicitly defined at any point during the Pswarm process. Both parameters can be deduced from a topographic map of the Pswarm projection and a dendrogram.

## 3.1    Projection with Pswarm

This section introduces the Polar swarm (Pswarm algorithm, which is the key foundation for the clustering performed in the DBS framework. Although the entire algorithm is used in an interactive clustering approach, Pswarm by itself may be used as a projection method.

In contrast to all other common projection methods [Venna/Kaski, 2007; Venna et al., 2010], Pswarm does not require any input parameters other than the data set of interest, in which case Euclidean distances are used in the input space. Alternatively, a user may also provide Pswarm with a matrix defined in terms of a particular dissimilarity measure, which is typically a distance but may also be a non-metric measure.

### 3.1.1    Motivation: Game Theory

The key idea of Pswarm is to redefine a game as one annealing step (epoch), the players as DataBots, and the scent as a payoff function and to find an equilibrium for each game. In the context of Pswarm, the game consists of rules governing the movement of the DataBots, which is defined by the grid, the neighborhoods and the payoff function. Each DataBot searches for its strongest payoff by either moving across the grid or staying in its current position. A new game (epoch), which is defined based on the considered neighborhood radius R, begins once an approximate equilibrium is achieved, i.e., once no movement of any DataBot leads to a stronger or better payoff for any other DataBot any longer (weak Nash equilibrium).

### 3.1.2    Symmetry Considerations

If we consider DataBots that occupy space in two dimensions, such as spheres or atoms, two points must be considered: first, no two DataBots are allowed to be in the same spot at the same time (*collision avoidance*), and second, a hexagonal lattice is the densest possible packing of identical spheres in two dimension**s** [Hunklinger, 2009, p. 65]. Every such sphere represents a possible position for a DataBot. To ensure that the two-dimensional output space is used most efficiently, a hexagonal lattice structure (*grid*) is used in Pswarm. To avoid problems associated with the surface of the grid, such as the positioning of DataBots near the border, the grid must have periodic boundary conditions and consequently must possess full *translational symmetry* [Haug/Koch, 2004, p. 34]. If the third dimension (e.g., as in a crystal) is disregarded, this two-dimensional grid can be represented by a three-dimensional torus [Pasquier, 1987], which is hereafter referred to as a *toroidal grid*. This means that the borders of the grid are cyclically connected. The periodicity of the grid is defined by its size in terms of the numbers of lines (*L*) and columns (*C*). If the grid were *planar* (not toroidal), undesired boundary effects could affect the outcome of any method.

Boundary effects are effects related to the borders of the output space in which the patterns of interactions across the borders of the bounded region are ignored or distorted, giving rise to shape effects, such that the shape imposed on the planar output space affects the perceived interactions between phenomena (see [McDonnell, 1995]). For example, if the output space is planar, it is unknown whether a projected point on the left border is similar (or dissimilar, in this case) to a projected point on the right border. It could be that the projection method is constrained to split similar points (with regard to the input space) in the output space. Another example is the distorted interactions between DataBots on the four borders when the output space is planar. Compared with a planar output space, a toroidal output space imposes fewer constraints on a projection (or clustering) method[5] and therefore enables a more optimized folding of the high-dimensional input space. A toroidal output space (in the case of Pswarm, a grid) possesses the advantage of translational symmetry in two dimensions, and in this case, the direction of a DataBot's movement is less important than its extent (length) because of the periodicity (of the grid).

In addition to the above considerations, the positions on the grid are coded using polar coordinates because the distances between DataBots on the grid will be most important in later computations of the neighborhoods and the annealing scheme. Consequently, based on the relevant *symmetry considerations,* a transformation of the Cartesian *(x, y)* coordinate system into polar coordinates $(r, \phi) \epsilon O$ is proposed as follows:

$$r = x^2 + y^2$$
$$\phi = tan^{-1}\left(\frac{y}{x}\right) * \frac{180}{\pi}$$

Hereafter, *r* represents the length of a DataBot's movement (*jump*), and $\phi$ represents the direction of that movement.

---

[5] To the author's knowledge, only the emergent self-organizing map (ESOM) and the swarm-organized projection (SOP) method offer the option to switch between planar and toroidal spaces (see [Ultsch, 1999], [Herrmann, 2011, p. 98]).

In Pswarm, the grid size is chosen automatically, subject to three conditions. Let $\tilde{D}$ be an upper triangle of the matrix of the input distances, let N be the number of DataBots, let $\alpha$ be the number of possible jump positions, let $\beta \in (0.5, 1]$ be a scaling factor, and let $\mathfrak{p}_{99}$ and $\mathfrak{p}_{01}$ denote the 99-th and first percentiles, respectively, of the distances; then, the conditions for determining the grid size are

$$\frac{\sqrt{C^2 + L^2}}{1} \geq \frac{\mathfrak{p}_{99}(\tilde{D})}{\mathfrak{p}_{01}(\tilde{D})} =: A$$

$$L * C \geq \alpha * N$$

$$\frac{L}{C} = \frac{\beta}{1}$$

These conditions result in the following bi-quadratic equation:

$$C^4 - A^2 * C^2 + \alpha^2 * N^2 = 0$$

$$z_{1/2} = A^2 \pm \frac{1}{2}\sqrt{A^4 - \frac{\alpha^2}{4}N^2}$$

$$\Rightarrow \quad C = \begin{cases} \frac{1}{\sqrt{2}}\sqrt{A^2 + \sqrt{A^4 - \frac{\alpha^2}{4}N^2}} & , if\ A^4 > \frac{\alpha^2}{4}N^2 \\ approximation, & otherwise \end{cases} \quad (3)$$

The first condition ensures that the shortest and longest distances of interest are assignable to grid units. It defines the possible resolution of high-dimensional structures in the grid. The second condition ensures that there are sufficient available positions to which a DataBot can jump. The third condition causes the grid to be more rectangular than square because in the case of SOMs, "rectangular maps outperform square maps" [Ultsch/Herrmann, 2005]. The first two conditions are used to formulate the bi-quadratic equation under the assumption of equality (see Eq. (3)). If the equation has no solution for the case of $A^4 < \frac{\alpha^2}{4}N^2$, then conditions I and III are used to generate approximate solutions. In this solution space, a solution that fulfills condition II is chosen by a heuristic optimization using a grid search. An example of a polar coordinate system with rectangular borders is drawn in Fig. 3.1.

### 3.1.3 Algorithm

Several previously developed ideas are applied in Pswarm: scent[6] [Herrmann/Ultsch, 2008a, 2009], DataBots [Ultsch, 2000b] and the decreasing neighborhood radius proposed for DataBots by [Kämpf/Ultsch, 2006]. The decrease in the radius is based on the data and is not predefined by parameters. The underlying idea of the decreasing radius approach is to promote self-organization, first of a global structure and then of local structures [Kämpf/Ultsch, 2006].

The intelligent agents of Pswarm operate on a toroidal grid where the positions are coded using polar coordinates, $i_\phi(r) \epsilon\, O$. This permits the DataBots' movement, the neighborhood function and the annealing scheme to be precisely defined. The numeric vector $z_j$ associated with each DataBot $b_j$ represents its distances from all other DataBots in the input space I.

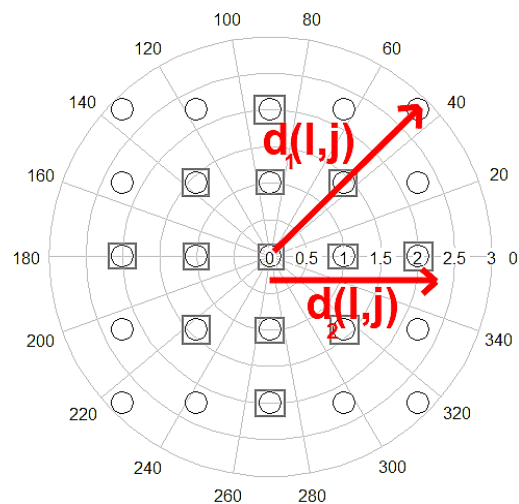

Figure 3.1: A rectangular lattice tiling of a square shape in polar coordinates. In Pswarm it applies $d_1(l,j) \neq d_2(l,j)$ for j, l= {$r, \phi$} in polar coordinates. All positions at distances smaller than or equal to $r = 2$ are marked by gray squares. In this case, the neighborhood (Eq. 4) depends on a precise one-dimensional grid distance. Independent of the coordinate system, the Pswarm (hexagonal) grid has a rectangular shape of borders, with a size of (*L, C*).

---

[6] Called topographic stress in [Herrmann/Ultsch, 2009].

The output-space distances are coded using only the polar coordinate *r*. The size of the squared-distance matrix D defines the number of DataBots *N*. After the assignment of initial random positions on the grid O (and therefore random output distances) to the DataBots in Listing 8.1, a data-driven decreasing of the radius R begins. In every iteration, a portion of the DataBots is allowed to jump if the payoff in one of their new positions is better (stronger) than that in their old positions. In other words, each DataBot is given a chance *c(R)* to try new positions on the grid.

The chance $c(R)$: $\mathbb{N} \rightarrow [0.05, 0.5]$ is a continuous, monotonically decreasing linear function addressing the number of the DataBots which are allowed to search for a new position to jump to[7]. Initially, many[8] DataBots are allowed to jump simultaneously to reproduce the coarse structure of the high-dimensional data set. However, as the algorithm progresses to address finer structures, only a small number[9] of DataBots may move simultaneously. The chance function depends on the number of DataBots and the current radius *R* and consequently is based on the data itself. In Pswarm, the length of a possible DataBot jump is not reduced during annealing[10].

---

[7] Details are unimportant, but can be inspected in the function Pswarm, line 145-147, 211 at in https://cran.r-project.org/package=DatabionicSwarm

[8] However, no more than half of the DataBots are allowed to search for a new position.

[9] At the end exactly five percent of all DataBots.

[10] Unlike in the SOP algorithm.

*function Positions O=Pswarm(matrix D(l, j))*

 *for all $z_i \in I$: assign an initial random polar position $i_\phi(r) \in O$ on the grid*

  *to generate DataBots $b_i \in B$*

 *for R={Rmax=Lines/2,...,Rmin} do*

  *calculate chance c(R)*

  **Repeat** *for each iteration*

   $c = sample(C(R), B),\ c \subseteq B$

   $m_k(c) = uniform(1, Rmax), with\ k = 1, \dots \alpha, m_k(c) \in O$

   $l(c) = \underset{j \in \{i, m_k(c)\}}{\operatorname{argmax}} (\lambda(b_j, R))$

   $l(!\,c) = i$

   $S = \sum_l\ \lambda_l(b_l, R)$

  **Until** $\dfrac{\partial S\big(e, \lambda(R)\big)}{\partial e} = 0$

 *return O in Cartesian coordinates*

 *end function Pswarm*

Listing 8.1: The Pswarm algorithm (details in Eq.4-6). New possible positions are depicted with $m_k(c)$ where k indicates up to the number of $\alpha$ possible jump positions chosen with an equal chance in the range from 1 up to Rmax relative to the old position $i_\phi(r)$ of each DataBot of the subset $c = \{b_1, \dots, b_j\}$ selected by chance C(R). After the decision to jump or not to jump derived using payoff with scent $\lambda$, the position is depicted with l(c). All other DataBots $!\,c = B \backslash c$ do not search for a new position and remain on their old position $i$.

The $\alpha$ possible jump positions of each sampled DataBot $b_{m \in \{1,\dots,j\}}$ to new positions $m_k(b_m)$ are drawn from a uniform distribution; therefore, the probability of selection is the same for all possible jumps, from a jump to zero to a jump to *Rmax* in any direction. The direction of a jump to a new position is chosen separately from among all positions corresponding to an equal jump length. This approach prevents local minima from causing the DataBots to become stuck in an incorrect cluster because the length of their jump is smaller than half of the cluster's diameter.

No DataBot is allowed to jump to an occupied position. Each DataBot may choose one of the four ($\alpha = 4$) best different positions in different directions to which to jump if it is sampled for jumping depicted by $c = \{b_1, \dots, b_j\} \subseteq B$. This approach ensures a high probability that every sampled DataBot will find a free position.

### 3.1.4 Data-based Annealing Scheme

Let each annealing step be defined as an epoch $e$; then, a new epoch begins (and a game[11] ends) if the radius $R$ is reduced by the condition defined below.

Let $r(j, l)$ be the one-dimensional distance from $l \in O$ to $j \in O$ in polar coordinates $(r, \phi)$ as specified by the radius $R_e$; then, the neighborhood function "Cone" is defined as

$$H_R : R_e \to [0, 1]:$$
$$H_R \begin{cases} 1 - \frac{r(j,l)^2}{\pi R_e^2}, & iff \ \frac{r(j,l)^2}{\pi R_e^2} < 1 \\ 0, & otherwise \end{cases} \qquad (4)$$

where $R_e$ is the radius of the neighborhood during epoch $e$.

Let D(l, j) be the distance between $x_l, x_j \in I$, and let $r(j, l)$ be the one-dimensional radial distance in two-dimensional polar coordinates $(r, \varphi)$ in the output space O; then, in Pswarm, the scent around a DataBot $b_j$ is redefined to

$$\lambda_e(b_j, R_e, S_0) = \begin{cases} S_0 - \frac{\sum_{l \in I} H_R(r(j,l)) * D(j,l)}{\sum_{l \in I} H_R(r(j,l))}, & iff \sum_{l \in W} H_R(r(j,l)) > 0 \\ S_0, & otherwise \end{cases} \qquad (5)$$

where $S_0 = \sum_j \left| \lambda \left( b_j, R_{max}, 0 \right) \right|$. Following the discussion in section 2, Eq. 1, the scent $\lambda \left( b_j, R \right)$ is identified in the payoff function $\lambda_e(b_j, R): \mathbb{R}_0^+ \times O \to \mathbb{R}_0^+$ with the specific neighborhood function for polar coordinates (Eq. 4) for a DataBot.

The high-dimensional input distances *D(l, j)* must be calculated only once, which is done prior to starting the algorithm, thereby reducing the computational cost. The computational cost of the algorithm does not depend on the dimension of the data set but does depend on the number of DataBots *N* and the number of possible jump positions $\alpha$ per DataBot. Additionally, Pswarm allows the conversion of distances or dissimilarities into two-dimensional points.

---

[11] In the context of game theory.

Let $e$ be the current epoch, let $R_e$ be the current neighborhood radius, and let $b_j \in B$ denote a DataBot; then, the sum of all payoffs is the current global happiness, which may be called the stress[12] $S(e)$, and is defined as

$$S(e, R_e) = \sum_j \lambda_e (b_j, R_e)$$

The neighborhood is reduced if the derivative of the current global happiness is equal to zero:

$$\frac{\partial S(e, R_e)}{\partial e} = 0 \qquad (6)$$

which is called the *equilibrium of happiness* condition. The neighborhood radius R is reduced from *Rmax* toward *Rmin* with a step size of 1 if the derivative of the sum of all payoffs $\lambda_e$ is equal to zero. This is the case if a (weak) equilibrium for all DataBots is found.

Because not all DataBots are allowed to jump simultaneously during a single iteration, as imposed by the function $sample(c(R), B)$, the DataBots are able to pay off their neighborhoods more often, thereby promoting the process of self-organization. By searching for an equilibrium, the net number of DataBots that would like to jump or are unhappy is irrelevant to the self-adaptive annealing process. Instead, the decision to shrink the neighborhood size or to proceed to the next epoch $e$ is made based on a Nash equilibrium [Nash, 1950]. The criterion is clearly defined to correspond to the condition in which the global amount of happiness in the current epoch remains unchanged, which is defined as the *equilibrium of happiness*, $\frac{\partial S}{\partial e} = 0$.

### 3.1.5 Annealing Interval

Rmax is equal to Lines/2 if Lines<Columns to prevent self-interaction of the DataBots. If the radius R were to be greater than Lines/2, then the neighborhood of a given DataBot would overlap with itself because of the toroidal nature of the grid. Moreover, the probability density function for choosing a new position cannot be uniformly (or Gaussian) distributed in this case because border positions can be reached from two directions $\phi$ on a toroidal grid.

Rmin is determined by the size of the grid and the number of DataBots. It is set to a value that allows every DataBot to smell a minimum of 5% of the other DataBots if they are distributed uniformly[13]. This selection is inspired by an emergent phenomenon called an ant mill [Schneirla, 1971, pp. 281-283]: Army ants are an aggressive, nomadic species, incessantly moving around. Based on its payoff, every ant follows another ant in front of it. If the head of the ant colony runs into the tail of the colony, the ants form a so-called circle of death, because they keep moving until they die. This phenomenon would not occur if the ants were able to smell a region farther ahead of them. In Pswarm this effect was observed in simulations on the dataset Hepta, where the clusters continuously moved from left to right on the toroidal plane.

---

[12] To simplify the comparison with SOP.

### 3.1.6   Convergence

In game theory, for a game with egoistic agents, a solution concept exists called the Nash equilibrium [Nash, 1950].

Let $(P, \Lambda)$ be a game with n DataBots $b_i, i = \{1, \dots, N\}$, where $P$ is a set of movement strategies and $\Lambda = \{\lambda_{e,i}(b_i, R_e = const) \mid i = 1, \dots, N\}$ is the payoff function evaluated for every grid position $w_i \in P_i$. Each DataBot chooses a movement strategy consisting of a probability associated with a position on the grid. Upon deciding on a position, a DataBot receives the payoff defined by the scent.  P is a set of mixed strategies that are chosen stochastically with fixed probability in the context of game theory. Nash proved that in this case, the following equilibrium exists:

$$\forall i, w_i, b_i \in P_i : \lambda_i \left( b_i^{'} \right) \geq \lambda_i \left( b_i \right) \qquad (7)$$

The strategy $b_i^{'}$ is the equilibrium, for which no deviation in strategy (position on the grid) by any single DataBot results in a greater profit for that DataBot. In the case of Pswarm, the Nash equilibrium in Eq. 7 is called weak because there may be more than one strategy with the same payoff for some DataBots. Because of the existence of this equilibrium, the Pswarm algorithm will always converge. After the algorithms converge in the final iteration, the coordinates are transformed back into the Cartesian system (see Listing 8.1).

### 3.2   Topographic Map and Clustering for Pswarm

The representation of the high-dimensional distance and density-based structures is defined by the projected points of the polar swarm (Pswarm) which is the first module of DBS. The projected points of Pswarm and their mapping to high-dimensional data points predefine the visualization of the topographic map. The topographic map has to be generated because Johnson–Lindenstrauss lemma [Dasgupta/Gupta, 2003] states that by limiting the Output space

---

[13] Rmin (and Rmax) are chosen automatically by the Pswarm algorithm based on the gird size and consequently based on the data.

to two dimensions, low dimensional similarities between projected points do not represent high-dimensional distances $D(I, j)$ coercively. The color scale of the topographic map is chosen to display various valleys, ridges and basins:

blue colors indicate small high-dimensional distances and high densities (sea level), green and brown colors indicate middle high-dimensional distances and densities (small hilly country) and shades of white colors indicate high distances and small densities (snow and ice of high mountains) [Thrun et al., 2016]. The valleys and basins indicate clusters and the watersheds of hills and mountains indicate borderlines of clusters [Thrun et al., 2016]. The color scale is combined with contour lines.

In brief, a topographic map with hypsometric tints is pictured which visualizes high-dimensional structures and can be 3D printed such that through its haptic form it is even more understandable for experts in the data's field [Thrun et al., 2016].

The idea to use this topographic map for an automatic clustering approach has a theoretical background ([Lötsch/Ultsch, 2014]). In brief, around each projected point a Voronoi cell can be computed [Toussaint, 1980]. The size of a cell is characterized in terms of the nearest data points surrounding the point assigned to that cell. Within the borders of one Voronoi cell, there is no position that is nearer to any outer data point than to the data point within the cell. Each two directly neighboring borders of two Voronoi cells define a vertex between two projected points. Thus, a Delaunay graph is derived. This Delaunay graph is weighted with the high-dimensional distances of the input space and can be used for hierarchical clustering if the right parameters are set [Ultsch et al., 2016a]. In this work, the shortest path between each two projected points is calculated out of the weighted Delaunay graph using the Djikstra algorithm [Dijkstra, 1959]. The shortest paths are used in a hierarchical clustering process which requires human intervention to choose the visualized cluster structure which is either connected or compact (see [Thrun, 2018] for details). Looking at the topographic map, this semi-interactive approach does not require any parameters other than the number of clusters and the cluster structure. If the number of clusters and the clustering method are chosen correctly, then the clusters will be well separated by mountains in the visualization. Outliers are represented as volcanoes and can be interactively marked in the visualization after the automated clustering process if they are not recognized sufficiently in the automatic clustering process. A simplified version of the self-organizing map (SOM) method applied in order to generate the visualization (topographic map) used in this work and an overview in SOMs is described in the co-submission for 'MethodsX'[Thrun/Ultsch, 2020b], and be downloaded from CRAN (https://cran.r-project.org/package=GeneralizedUmatrix).

For example, the topographic map in Fig. 3.1 depicts two well-separated clusters (green and blue), which the compact DBS clustering can detect. The outliers in a data set may be manually identified by the user. In the other case, choosing the connected structure option for the clustering process would result in the automatic detection of all outliers. However, this option does not always lead to the detection of the main clusters in terms of the shortest path distances.

## 4    Methods

This section describes the parameter settings for the various clustering algorithms. All the data sets used in the results section are described separately in the co-submission for 'Data in Brief': "Clustering Benchmark Datasets Exploiting the Fundamental Clustering Problems"[Thrun/Ultsch, 2020a]. Patient consent was obtianed for the leukemia dataset, in accordance with the Declaration of Helsinki, and the Marburg local ethics board approved the study (No. 138/16).

### 4.1    Parameter Settings

For the k-means algorithm, the CRAN R package cclust was used (https://cran.r-project.org/web/packages/cclust/index.html). The k-means algorithm was restarted for each iteration from random starting points. For the single linkage (SL) and Ward algorithms, the CRAN R package stats was used (https://cran.r-project.org/web/packages/stats/index.html). For the Ward algorithm, the option "ward.D2" was used, which is an implementation of the algorithm as described in [Ward Jr, 1963]. For the spectral clustering algorithm, the CRAN R package kernlab was used (https://cran.r-project.org/web/packages/kernlab/index.html) with the default parameter settings: "The default character string "automatic" uses a heuristic to determine a suitable value for the width parameter of the RBF kernel", which is a "radial basis kernel function of the "Gaussian" type". The "Nyström method of calculating eigenvectors" was not used (=FALSE). The "proportion of data to use when estimating sigma" was set to the default value of 0.75, and the maximum number of iterations was restricted to 200 because of the algorithm's long computation time (on the order of days) for 100 trials using the FCPS data sets. For the mixture of Gaussians (MoG) algorithm, the CRAN R package mclust was used (https://cran.r-project.org/web/packages/mclust/index.html). In this instance, the default settings for the function "Mclust()" were used, which are not specified in the documentation. For the partitioning around medoids (PAM) algorithm, the CRAN R package cluster was used (https://cran.r-project.org/web/packages/cluster/index.html). The clustering of DBS always provides a fixed number of clusters, once it is set. DBS is available as the R package "DatabionicSwarm" on CRAN.

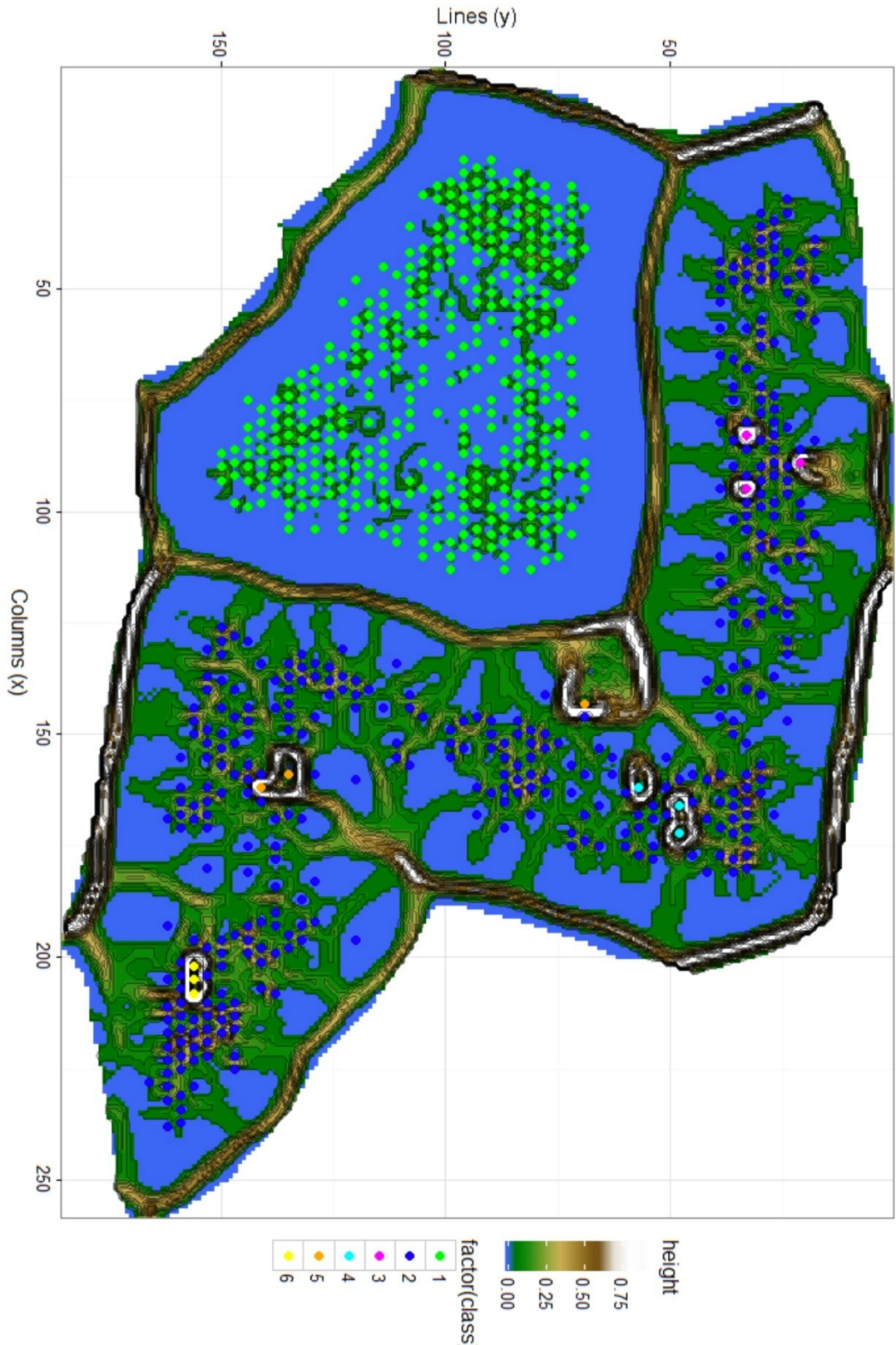

Figure 3.1: Top view of the Topographic map of the Target data set [Thrun/Ultsch, 2020a]. Two main clusters are shown; the cluster labeled in green has a higher density than the cluster labeled in blue. The outliers (orange, yellow, magenta and cyan) lie in volcanoes.

## 5    Results on Pre-classified Data Sets

The figures 5.1, 5.2, and 5.3 show the performance of several common clustering algorithms[14] compared with DBS based on 100 trials. In Figure 5.1 the performance is depicted using common boxplots of the error rate for which 50% is the level attributable to chance. In Fig. 5.2 the probability density function of each error rate is estimated by the Mirrored-Density plot (MD plot) enabling the visualization of skewness and multimodalit [Thrun et al., 2020]. In Figure 5.3, the F measure with beta=1 (F1-score) ([Chinchor, 1992] cites [Van Rijsbergen, 1979]) is visualized with the MD plot. All FCPS data sets have uniquely unambiguously defined class labels. For the error rate is defined as 1-Accuracy was used as a sum over all true positive labeled data points by the clustering algorithm. The best of all permutation of labels of the clustering algorithm regarding the accuracy was chosen in every trial because the labels are arbitrarily defined by the algorithms.

Here, the common clustering algorithms considered are single linkage (SL) [Florek et al., 1951], spectral clustering [Ng et al., 2002], the Ward algorithm [Ward Jr, 1963], the Linde-Buzo-Gray algorithm (LBG-k-means) [Linde et al., 1980], partitioning around medoids (PAM) [Kaufman/Rousseeuw, 1990] and the mixture of Gaussians (MoG) method[15] with expectation maximization (EM) [Fraley/Raftery, 2002]. Aside from the number of clusters, which is given for each of the artificial FCPS data sets, only the default parameter settings of the clustering algorithms were used. ESOM/U-matrix clustering [Ultsch et al., 2016a] and DBscan [Ester et al., 1996] were omitted because no default clustering settings exist for these methods. K-means has the highest overall error rate, and spectral clustering shows the highest variance. The results for the other clustering algorithms vary depending on the data set. DBS has the lowest overall error rate. However, on the Tetra data set, it is outperformed by PAM and MoG; on the EngyTime data set, it is outperformed by MoG; and in the case of the WingNut data set, it is outperformed by spectral clustering.

---

[14] They were chosen for the reason that they are freely available as CRAN packages and have a default setting of parameters besides the number of clusters and therefore a clustering can be automatically performed by these methods.

[15] "Clustering via mixtures of parametric probability models is sometimes in the literature referred to as 'model-based clustering'" [C. Hennig, et al. (Hg.), 2015, p. 10]. With the clustering algorithm of [Fraley/Raftery, 2006] in mind, here, this clustering method is called the *mixture of Gaussians* (MoG) method.

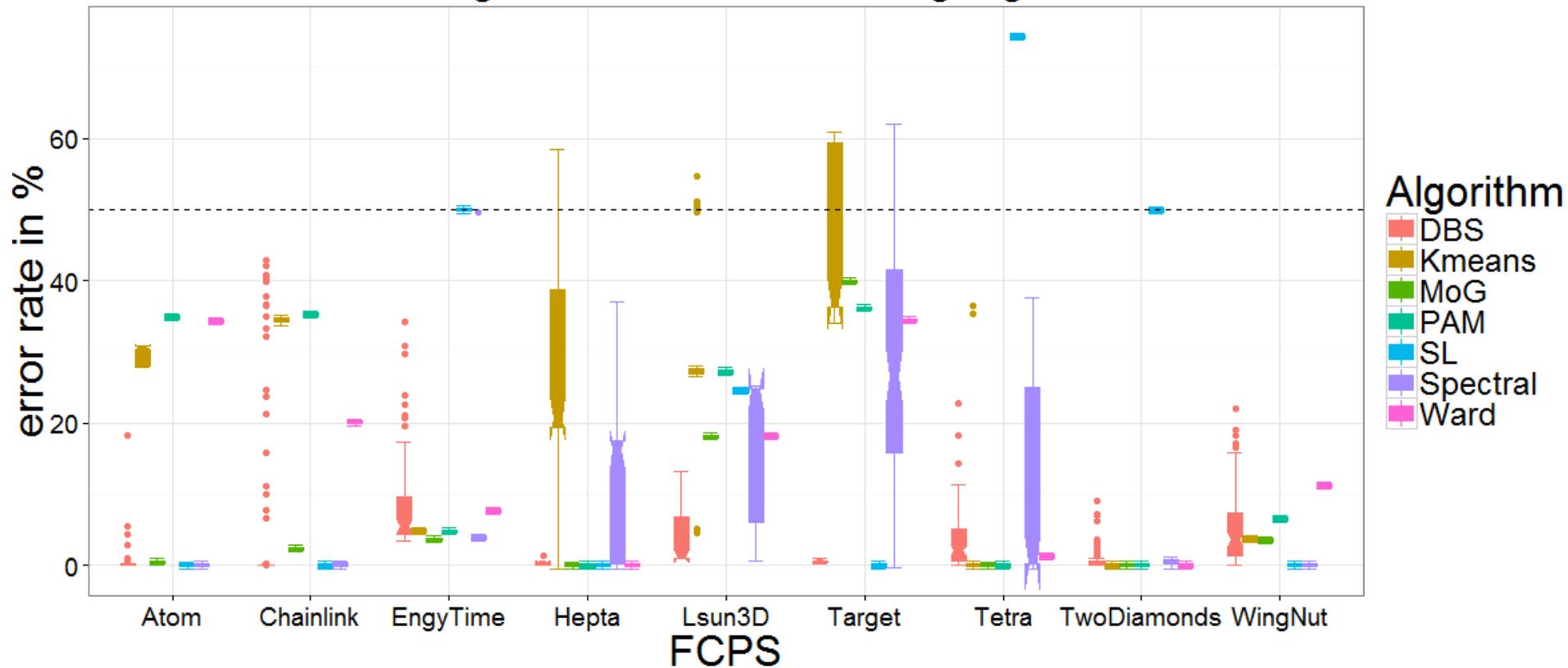

Figure 5.1: The results of 100 trials of common clustering algorithms on nine FCPS data sets, shown as boxplots. The dashed line at 50% represents the level of error attributable to chance. The notch in each boxplot indicates the median. The interactive clustering approach of DBS was not used here.

Abbreviations: median (notch), single linkage (SL), Linde-Buzo-Gray algorithm (LBG-k-means), partitioning around medoids (PAM), mixture-of-Gaussians clustering (MoG), Databionic swarm (DBS).

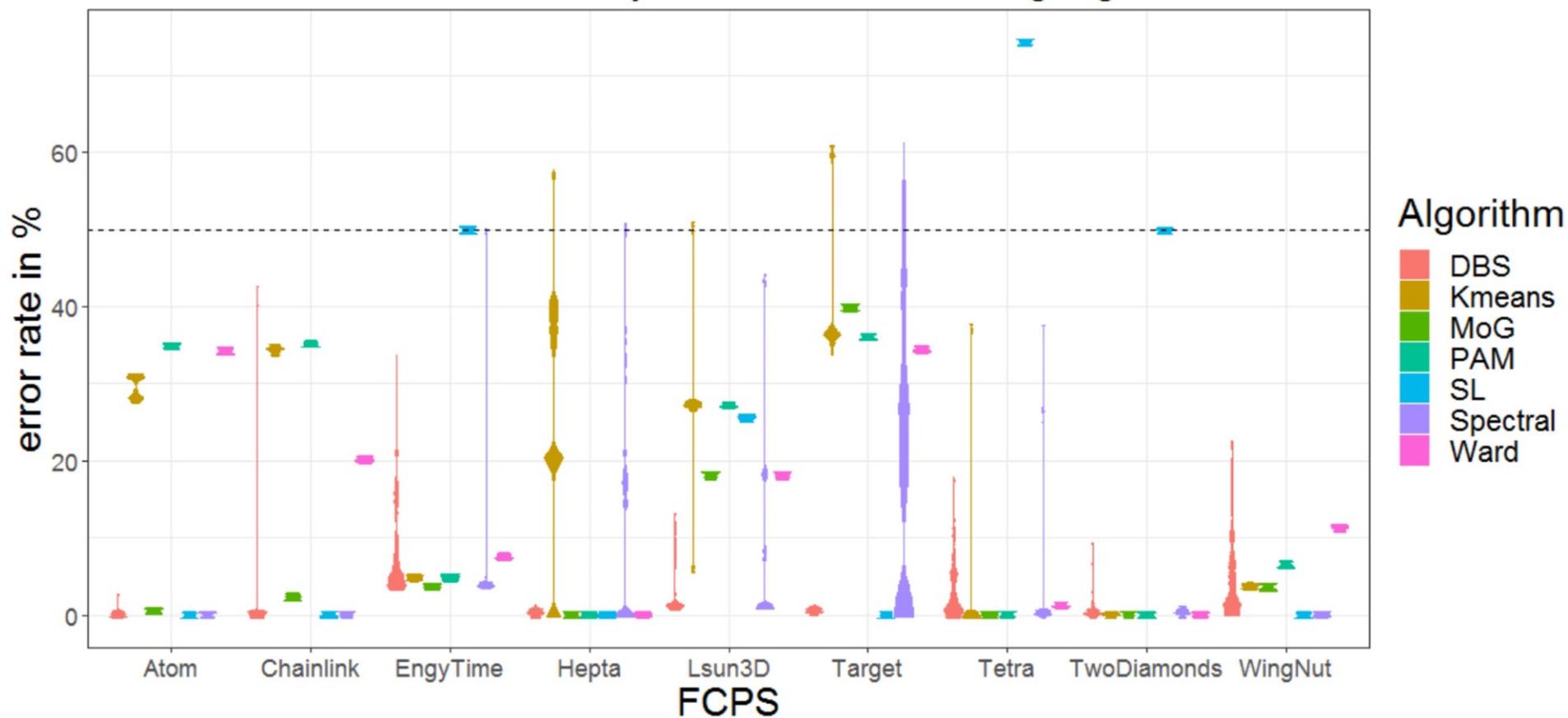

Figure 5.2 The results of 100 trials of common clustering algorithms on nine FCPS data sets, shown as Mirrored Density plots (MD plots)[Thrun et al., 2020]. The dashed line at 50% represents the level of error attributable to chance. The width of the violin (rotated density plot on each side) defines the error probability by estimating the empirical probability density function. The interactive clustering approach of DBS was not used here.

Abbreviations: median (notch), single linkage (SL), Linde-Buzo-Gray algorithm (LBG-k-means), partitioning around medoids (PAM), mixture-of-Gaussians clustering (MoG), Databionic swarm (DBS).

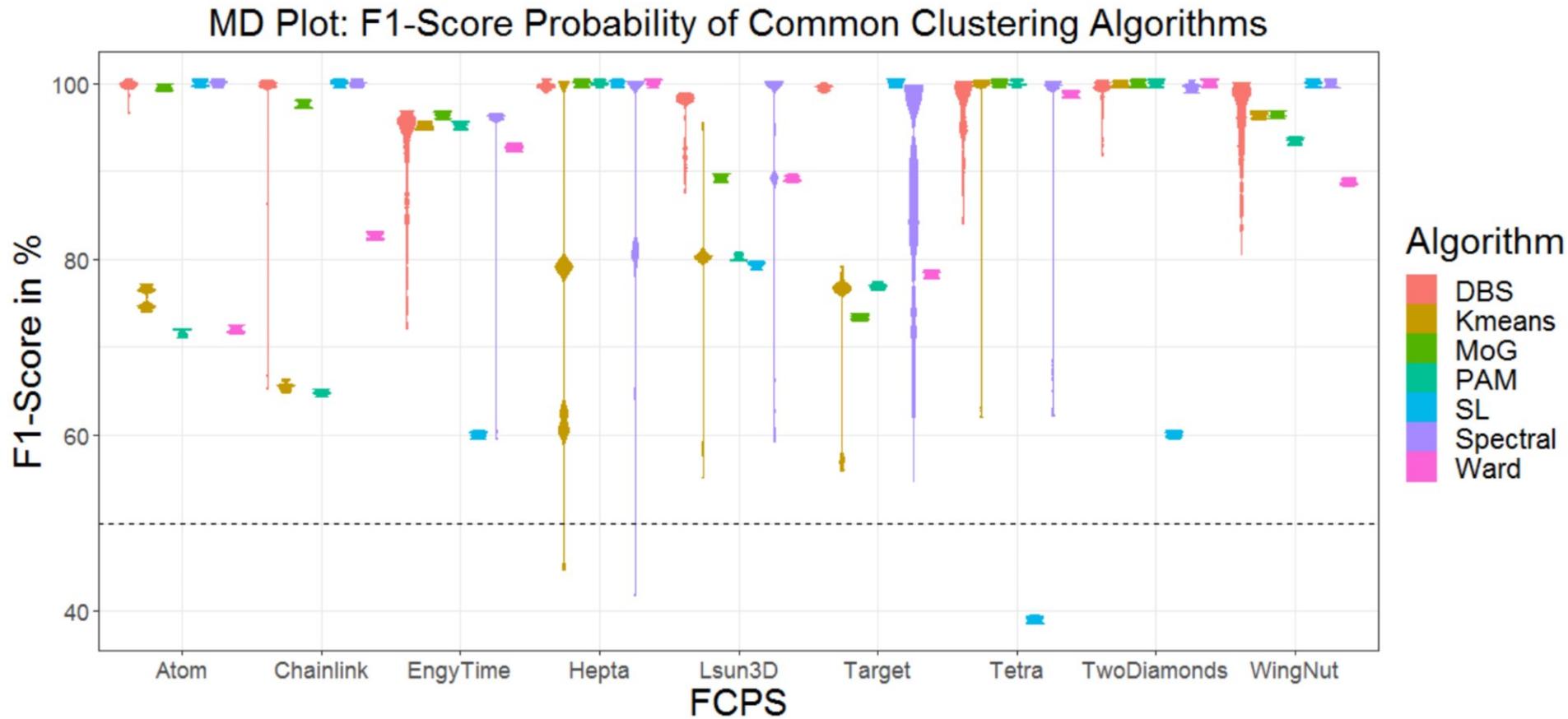

Figure 5.3 The results of 100 trials of common clustering algorithms on nine FCPS data sets, shown as Mirrored Density plots (MD plots)[Thrun et al., 2020]. The width of the violin (rotated density plot on each side) defines the error probability by estimating the empirical probability density function. The interactive clustering approach of DBS was not used here. The F1-Score is defined as the F measure with equal weighting of precision and recall.

Abbreviations: median (notch), single linkage (SL), Linde-Buzo-Gray algorithm (LBG-k-means), partitioning around medoids (PAM), mixture-of-Gaussians clustering (MoG), Databionic swarm (DBS).

### 5.1    Recognition of the Absence of Clusters

The Golf Ball data set does not exhibit natural clusters. Therefore, it is analyzed separately because, with the exception of SL and the Ward algorithm, the common clustering algorithms give no indication regarding the existence of clusters. This "cluster tendency problem has not received a great deal of attention but is certainly an important problem" [Jain/Dubes, 1988, p. 222]. The Ward algorithm indicates six clusters, whereas SL indicates two clusters (Figure 5.). As seen from the two dendrograms generated using DBS, the connected approach does not indicate any clusters, whereas the compact approach indicates four clusters (Figure 5.). However, the presence of four clusters is not confirmed by the DBS visualization:

In Figure 5.4, the DBS visualization of a topographic map is shown. The topographic map does not indicate a cluster structure. The compact DBS clustering divides the data points lying in valleys into different clusters and merges the data points into clusters through hills, resulting in cluster borders that are not defined by mountains.

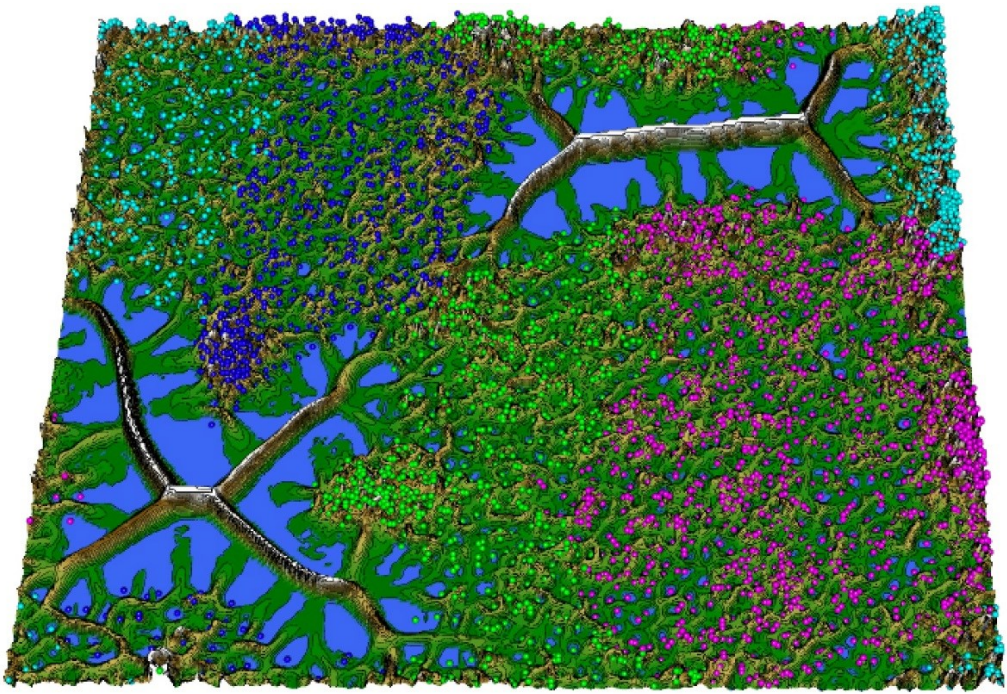

Figure 5.4: The topographic map of the structures of the Golfball dataset where the DBS projection and (compact) clustering are included. The visualization does not indicate a cluster structure because no valleys are visible and the mountain ranges are not closed. The DBS clustering generates clusters that are not separated by mountains. The visualization is toroid, i.e left-right and top-bottom borders are identical. However, no closed island can be extracted.

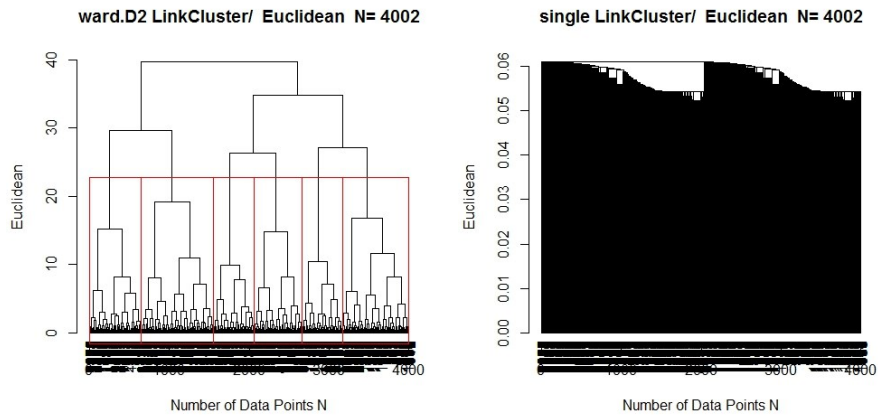

Figure 5.5: The dendrogram generated using the Ward algorithm indicates at least two clusters (top) with a high intercluster distance. The SL dendrogram indicates two clusters with a low intercluster distance.

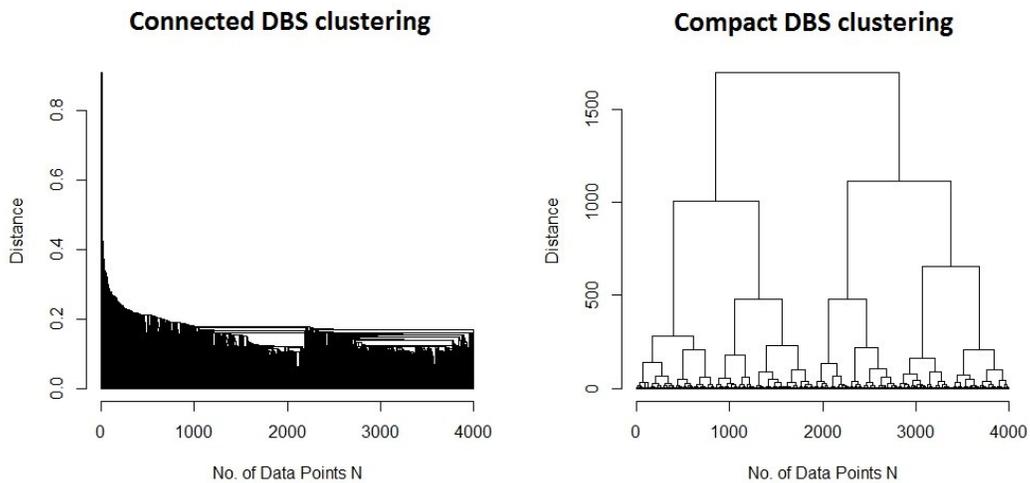

Figure 5.6: The two dendrograms generated using DBS. The connected approach does not indicate any clusters, whereas the compact approach does indicate four clusters. However, Figure 5. shows that these clusters are inconsistent with the visualization.

# 6    DBS on Natural Data Sets

Two real-world data sets are exemplarily used in this section to show that Databionic swarm (DBS) is able to find clusters in a variety of cases. Additional examples showing the reproduction of high-dimensional structures are shown in the SI B for pain genes [Thrun, 2018, p. 145] in Fig. B.2, for noxious cold stimuli [Weyer-Menkhoff et al., 2018] in Fig. B.3 and the Swiss banknotes [Flury/Riedwyl, 1988] in Fig. B.4. Using high-dimensional structures of clusters new knowledge was generated in the case of the world gross domestic product [Thrun, 2019] visualized in Fig B.5, in hydrology [Thrun et al., 2018] shown in Fig. B.6 and for of next-generation sequencing regarding the *TRPA1/TRPV1* NGS genotypes [Kringel et al., 2018] depicted in Fig. B.7. Additional examples can be found for the reproduction of classifications of common data sets like Iris [Anderson, 1935], and the Wine data set [Aeberhard et al., 1992]  (see  [Thrun, 2018, p. 187 ff]).

## 6.1    Types of Leukemia

The leukemia data set consists of 7747 variables for 554 subjects. Of the subjects, 109 were healthy, 15 were diagnosed with acute promyelocytic leukemia (APL), 266 had chronic lymphocytic leukemia (CLL), and 164 had acute myeloid leukemia (AML). The leukemia data set is a high-dimensional data set with natural clusters specified by the illness status.

Figure 6.1 shows the topographic map of the healthy patients and the patients diagnosed with these three major types of leukemia. The four groups are well separated by mountains, with the subjects represented by points of different colors. Magenta points indicate healthy subjects, whereas points of other colors indicate ill subjects. The automatic clustering of DBS is able to separate the four groups with an accuracy of 99.6%. Out of the six methods for conventional algorithms, only Ward is able to reproduce the four main clusters (SI A, Figure A.1).

Two outliers can be seen in Figure 6.1, marked with red arrows and lie in a volcano and on top of a white hill. These green and yellow outliers cannot be explained without de-anonymization of the patients, which was not feasible for the authors.

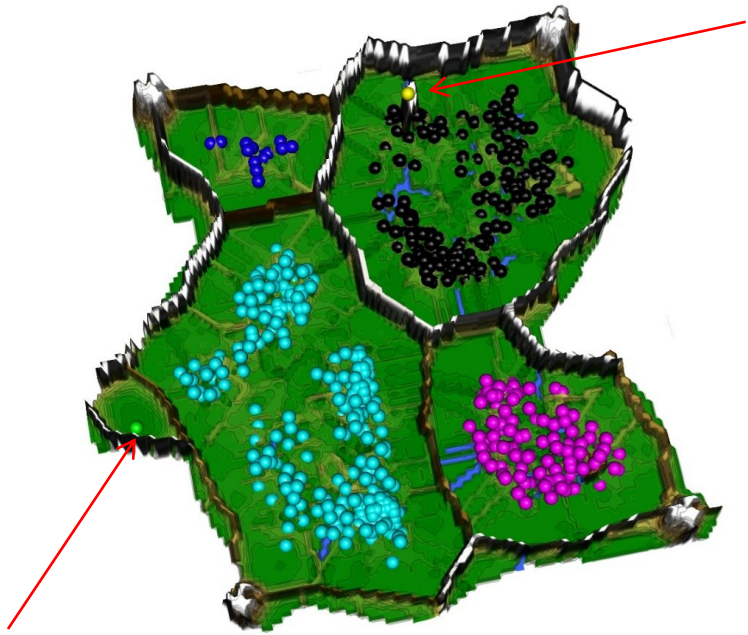

Figure 6.1: Topographic map of the high-dimensional structures of Leukemia shows the four main clusters separated by brown and white mountain ranges lying in green valleys. DBS clustering compared to the prior classification of the leukemia data resulted in an accuracy of 99.6%.. Top: healthy (magenta), Two outliers are marked with red arrows: an APL outlier (green) and a CLL outlier (yellow). The other clusters are:healthy (magenta), AML (cyan), APL (blue), and CLL (black).

## 6.2    Tetragonula Bees

The Tetragonula data set was published in [Franck et al., 2004] and contains the genetic data of 236 Tetragonula bees from Australia and Southeast Asia, expressed using 13 variables with a specific distance definition (see Data in Brief article "Clustering Benchmark Datasets Exploiting the Fundamental Clustering Problems" for details). The heatmap of the distances is shown in Figure 6.2**.**

The topographic map in Fig. 6.8 shows eight clusters and three types of Outliers. The silhouette plot indicates a hyperspherical cluster structure **(**Figure 6.3**)**. Additionally, using the prabclus package, the largest within-cluster gap, the cluster separation, and the average within-cluster dissimilarity of [C. Hennig, 2014] were calculated to be 0.5, 0.33 and 0.29, respectively. These values are the minima reported in [C. Hennig, 2014], presented there in Figure 4. Seven clusters of the average linkage hierarchical clustering with ten clusters ([C. Hennig, 2014, p. 5]) could be reproduced  (see Table C.1, SI C) with a total accuracy of 93%. Finally, as Fig.6.5 shows, the clusters strongly depend on the geographic origins of the bees as illustrated in [Franck et al., 2004, p. 2319].

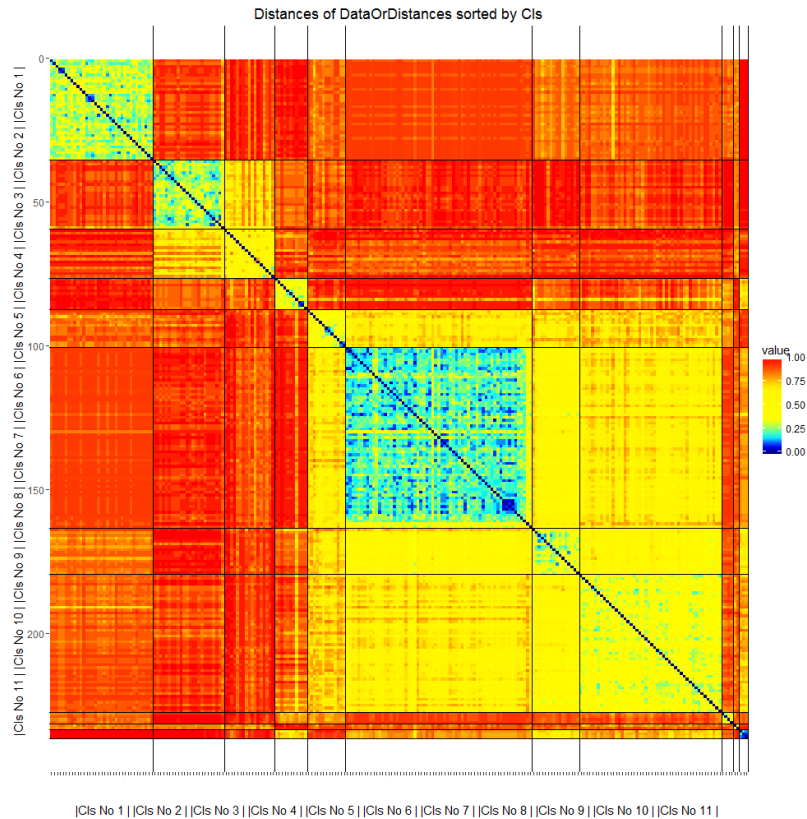

Figure 6.2: Heatmap of the distances for the Tetragonula data set.

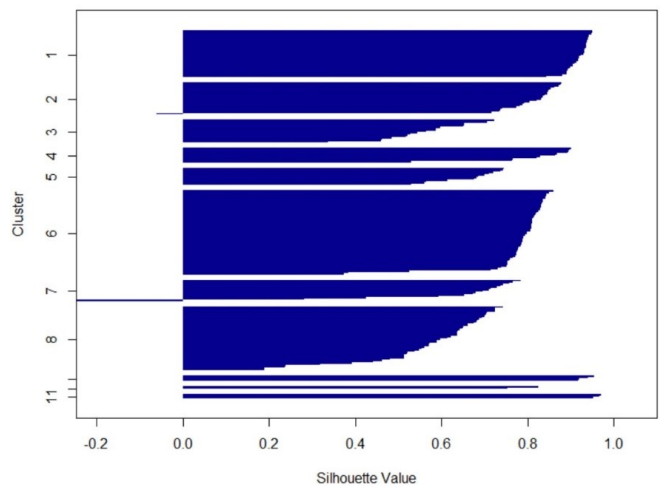

Figure 6.3: Silhouette plot of the Tetragonula data set, showing very homogeneous cluster structures.

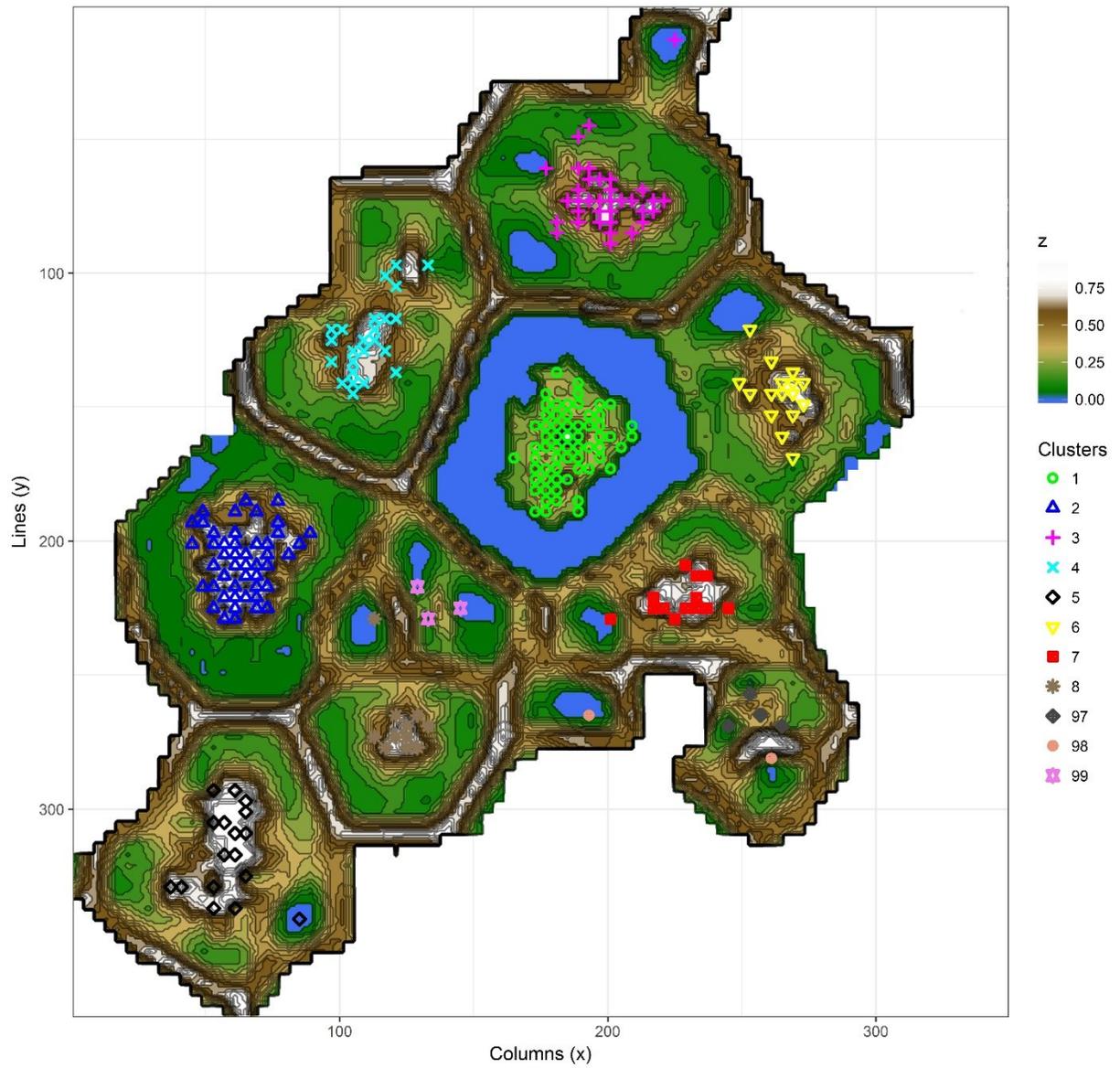

Figure 6.4: Top view of the topographic map of the high-dimensional structures of the Tetragonula data set. The clusters lie in green valleys and are separated by brown and white mountains. The cluster labels are colored as shown on the right. Similar color code is used in Fig.6.5 below. Cluster numbers are assigned in ascending order of prevalence.

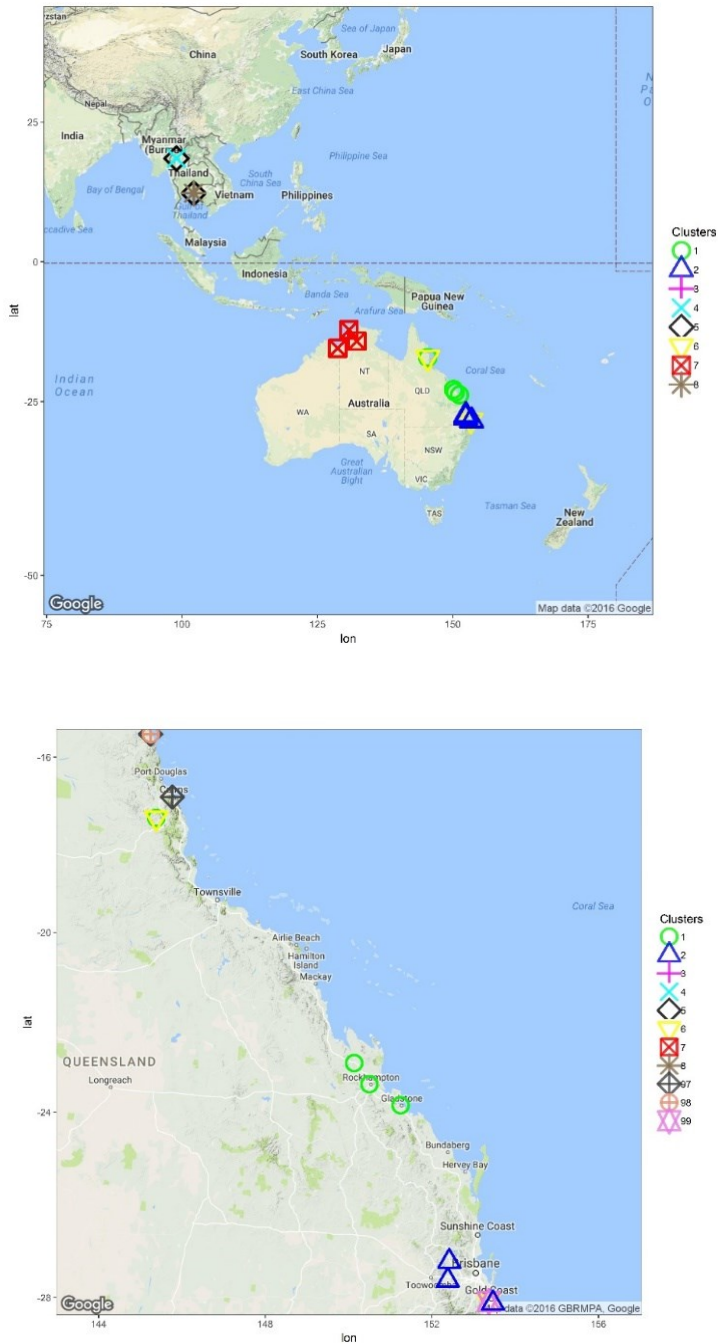

Fig.6.5: Clustering of the Tetragonula dataset is consistent with the geographic origins. **"Longitude and latitude of locations of individuals in decimal format, i.e. one number is latitude (negative values are South), with minutes and seconds converted to fractions. The other number is longitude (negative values are West)"** (see [C. Hennig, 2014] and the prabclus package). After the transformation into a two-dimensional plane, the first eight clusters (96% of data) are consistent with the geography (top) except for the Outliers in Queensland (bottom).

## 7    Discussion

A clustering approach "must be adaptive or exhibit 'plasticity,' possibly allowing for the creation of new clusters, if the data warrants it. On the other hand, if the cluster structures are unstable […], then it is difficult to ascribe much significance to any particular clustering. This general problem has been called 'the stability/plasticity dilemma' " [Duda et al., 2001].

The results of automatic DBS clustering with no user intervention were compared with the results of the common clustering algorithms k-means [MacQueen, 1967], partitioning around medoids (PAM) [Kaufman/Rousseeuw, 1990], single linkage (SL) [Florek et al., 1951] and spectral clustering [Ng et al., 2002], the mixture of Gaussians (MoG) method [Fraley/Raftery, 2002] and the Ward algorithm [Ward Jr, 1963]. It should be noted, that several of the comparative studies […] conclude that Ward's method […] outperforms other hierarchical clustering methods" [Jain/Dubes, 1988, p. 81]. MoG clustering, which is also known as model-based clustering, serves as the reference technique [Bouveyron/Brunet-Saumard, 2014]. Benchmarking with nine difficult datasets shows, that, compared to other algorithms, DBS is more likely to reproduce density and distance based structures with and without outliers.

In general, no algorithm can exist which outperforms every other algorithm in every case [Wolpert, 1996]. In this work, the main limitation of DBS is its complexity and variance; more than 4000 cases would require more than one day to compute because every case has to be represented in one DataBot and so far only one core in a PC can be used.

The large variance is mandatory: In error theory, the error of an algorithm is the sum of the components bias, variance and data noise (c.f. [Geman et al., 1992; Gigerenzer/Brighton, 2009]) Here, the bias is the difference between the cluster structures given and the ability to reproduce these structures. Variance is the stochastic property of reproducing the same result in different trials. Outliers in these data sets represent the data noise.

The results clearly show that, usually, either the variance is small and the bias large (single linkage, ward, PAM, model-based clustering) or the variance is large and the bias small (Spectral, DBS). The figures also show that data noise represented by outliers has a high influence on spectral clustering (Lsun3D, Target). The exception of this theory is found in k-means, which has often a high variance and a high bias.

The advantage of DBS is that besides the clustering a visualization of a topographic map is given. Looking into the visualization one would have the chance to intercept incorrect clustering results and perform another trial until visualization and clustering overlap resulting in a significant reduction of the variance.

The benchmarking on the FCPS datasets outline that it is preferable not to use a global objective function for clustering if no prior knowledge about data structures is used. For example, the k-means clustering algorithm imposes a spherical cluster structure [Duda et al., 2001, p. 542; Handl et al., 2005, p. 3202; Mirkin, 2005, p. 108; Theodoridis/Koutroumbas, 2009, p. 742; C. Hennig, et al. (Hg.), 2015, p. 61] such that the clusters cannot be too elongated [Kaufman/Rousseeuw, 2005, p. 117] and is sensitive to noise and outliers [Theodoridis/Koutroumbas, 2009, p. 744]. Therefore, k-means fails on Lsun3D (one of the clusters is elongated), Chainlink and Atom (non-spherical clusters) and Target (Outliers). The same argumentation is valid for PAM.

Single linkage (SL) searches for nearest neighbors [Cormack, 1971, p. 331], it tends to produce connected and chain-like structures [Hartigan, 1981; Jain/Dubes, 1988, pp. 64-65; Duda et al., 2001, p. 554; Everitt et al., 2001, p. 67; Theodoridis/Koutroumbas, 2009, p. 660]. Thus, SL fails for the datasets Lsun3D (one of the clusters is spherical), Tetra and TwoDiamonds (compact clusters touching each other) and EngyTime (overlapping clusters of varying density).

The Ward algorithm is sensitive to outliers and tends to find compact clusters of equal size [Everitt et al., 2001, p. 61, Tab. 1] that have an ellipsoidal structure [Ultsch/Lötsch, 2017]. Thus, Ward fails to cluster Atom and Chainlink (non-compact clusters), Lsun3D (one of the clusters is elongated), Target (outliers). Additional Ward seems to have difficulties with changing density (WingNut). The clustering criterions of Spectral clustering and Model-based clustering (MoG) are not that well understood. In literature the opinions are contradictory:

"They [spectral clustering methods] are well-suited for the detection of arbitrarily shaped clusters but can lack robustness when there is little spatial separation between the clusters" [Handl et al., 2005, p. 3202]. Spectral clustering is based on graph theory and consequently searches for connected structures [Ng et al., 2002, p. 5] of clusters with "chain-like or other intricate structures" [Duda et al., 2001, p. 582]. The FCPS datasets indicate that spectral clustering is unable to cope with outliers (Target, Lsun3D), spherical clusters of varying density (Hepta), and Tetra(compact clusters touching each other).

Jains and Dubes report that "fitting a mixture density model to patterns creates clusters with hyper-ellipsoidal shapes [Jain/Dubes, 1988, p. 92]. [Handl et al.] report that the MoG method is very effective for well-separated clusters [Handl et al., 2005, p. 3202]. Our trials cannot confirm that MoG is always effective for well-separated clusters (Lsun3D) and the method has difficulties with outliers (Target). Additionally, MoG method suffers "from the well-known curse of dimensionality [Bellman, 1957], which is mainly due to the fact that model-based clustering methods are over-parametrized in high-dimensional spaces" [Bouveyron/Brunet-Saumard, 2014, p. 53]. Thus, this approach cannot compute a result for Leukemia (Fig A.1).

For a clustering algorithm, it is relevant to test for the absence of a cluster structure [Everitt et al., 2001, p. 180], or the clustering tendency [Theodoridis/Koutroumbas, 2009, p. 896]. Usually, tests for the clustering tendency rely on statistical tests [Theodoridis/Koutroumbas, 2009, p. 896]. Unlike other hierarchical clustering algorithms (except for ESOM/U-matrix clustering [Ultsch et al., 2016a]), the DBS algorithm finds no clusters if no natural clusters exist (see also [Thrun, 2018, p. 192]). The clustering tendency is visualized by topographic map based on the generalized U-matrix. Additionally, the ESOM algorithm is to the knowledge of the authors the only algorithm where it was mathematically proven that no objective function exists [Erwin et al., 1992]. However, DBS was not compared to ESOM/U-matrix clustering because in [Ultsch et al., 2016a] the parameters were carefully chosen by the authors which do not allow for an automatic clustering approach. In general, the ESOM/U-matrix approach is only used for the visualization of structures.

In terms of stability and plasticity (Figure 5. and Figure 5.,), the Databionic swarm (DBS) framework outperforms common algorithms in clustering tasks on the FCPS. "One source of this dilemma is that with clustering based on a global criterion, every sample can have an influence on the location of a cluster center, regardless of how remote it might be" [Duda et al., 2001]. In contrast to standard approaches, swarm techniques are known for their properties of flexibility and robustness [Bonabeau/Meyer, 2001; Şahin, 2004].

Up to this point, mainly artificial datasets have been used to assess the capabilities of DBS. However, the introduction of a new clustering method is necessary only if it is useful. Therefore, two complex real-world data sets were analyzed using DBS to confirm its ability to reproduce known knowledge. The silhouette plots and the heatmaps, which showed small intracluster distances and large intercluster distances, indicated that the clustering results for both data sets were valid. On the high-dimensional leukemia dataset DBS outperformed conventional clustering algorithms with the exception of Ward in 10 trials (Fig. A.1) and preserves the high-dimensional structures given by the illnesses. In a direct comparison to a prior clustering DBS reproduced the result with an accuracy of 93% in Tetrangula dataset. It improves the prior clustering by detecting outliers more robustly. The clusterings of DBS are consistent with geographic information (Fig.6.9, A.6.) because they are consistently in one geographic region and all outliers lie in Queensland for Tetrangula. For the gross-domestic-product of 160 countries, the first cluster consists of mostly

African and Asian countries and a second cluster consists of mostly European and American countries [Thrun, 2019]. Different types of distance and density-based high-dimensional structures are shown in Fig. A2, A.3 and A.4 where DBS is able to preserve them. Using DBS, new knowledge was extracted out of high-dimensional structures with regards to genetic data and multivariate time series (Fig. A5, A.7).

In sum, the results indicate the main advantage of DBS which is its robustness regarding very different types of distance and density-based structures of clusters in natural as well as artificial datasets (Fig. 5.1). As a swarm technique, DBS clustering is more robust with respect to outliers than conventional algorithms which was shown on the artificial datasets of Lsun3D and Target, as well as the complex real-world datasets of Tretangula (Table A.1) and Leukemia (Fig. A.1).

## 8    Conclusion

A new approach for cluster analysis is introduced in this work. A theoretical comparison of ant-based clustering against SOM described in section 2 indicates that DBS clustering preserves high-dimensional structures (see p. 7, Eq. 1 and p. 17, Eq. 5). The benchmarking on artificial datasets supports this assumption. DBS is a flexible and robust clustering framework that consists of three independent modules. The first module is the parameter-free projection method Pswarm, which exploits the concepts of self-organization and emergence, game theory, swarm intelligence and symmetry considerations. The second module is a parameter-free high-dimensional data visualization technique, which generates projected points on a topographic map with hypsometric colors [Thrun et al., 2016] based on the generalized U-matrix [Ultsch/Thrun, 2017]. The third module is the clustering method itself with non-critical parameters. The clustering can be verified by the visualization and vice versa. The term DBS refers to the method as a whole. DBS enables even a non-professional in the field of data mining to apply its algorithms for visualization and/or clustering to data sets with completely different structures drawn from diverse research fields, simply by downloading the corresponding R package on CRAN [Thrun, 2017].

If prior knowledge of the data set to be analyzed is available, then a clustering algorithm and parameter setting that is appropriately chosen with regard to the structures that should be preserved can outperform DBS (Fig. 5.1). The general claim made in literature can be reproduced by this paper: "[T]he majority of clustering algorithms [...] impose a clustering structure on the data set X, even though X may not possess such a structure" [Theodoridis/Koutroumbas, 2009, p. 863]. Additionally, they may return meaningless results in the absence of natural clusters [Cormack, 1971, pp. 345-346; Jain/Dubes, 1988, p. 75; Handl et al., 2005, p. 3203]. The results presented in this work illustrate that the DBS algorithm does not suffer from these two disadvantages.

In this work, we give an overview over swarm intelligence and self-organization and use particular definitions for swarms and emergence which are based on an extensive review of literature. One of the contributions of this work is the outline the missing links between swarm-based algorithms and emergence as well as game theory. Further research is necessary on the application of game theory as the foundation for a data-based annealing scheme. At this point, it can be proven only that a weak Nash equilibrium will be found [Nash, 1951].

If the particular definitions for the concepts are applied, then, to the author's knowledge, DBS is the first swarm-based technique showing emergent properties while simultaneously exploiting the concepts of swarm intelligence, self-organization and the Nash equilibrium concept from game theory. The specific application of these concepts results in the elimination of a global objective function and eliminates the need to set ominous and yet critical parameters for clustering Algorithms.

## Acknowledgments

Special acknowledgment goes to PD. Dr. Cornelia Brendel of Univ. Marburg, Prof. Andreas Neubauer of Univ. Marburg, and Prof. Torsten Haferlach, MLL (Münchner Leukämielabor) for data acquisition and provision of the leukemia dataset.

## Supplementary Information A: Reproducibility of Structures based on Diagnosis of Leukemia for Common Clustering Algorithms

The interactive clustering approach of DBS was not used in Figure A.1. As the other algorithms are not robust against outliers, the number of clusters was set to four if required by the algorithm. Results for MoG (model-based clustering) were not computable due to the high-dimensionality of the dataset.

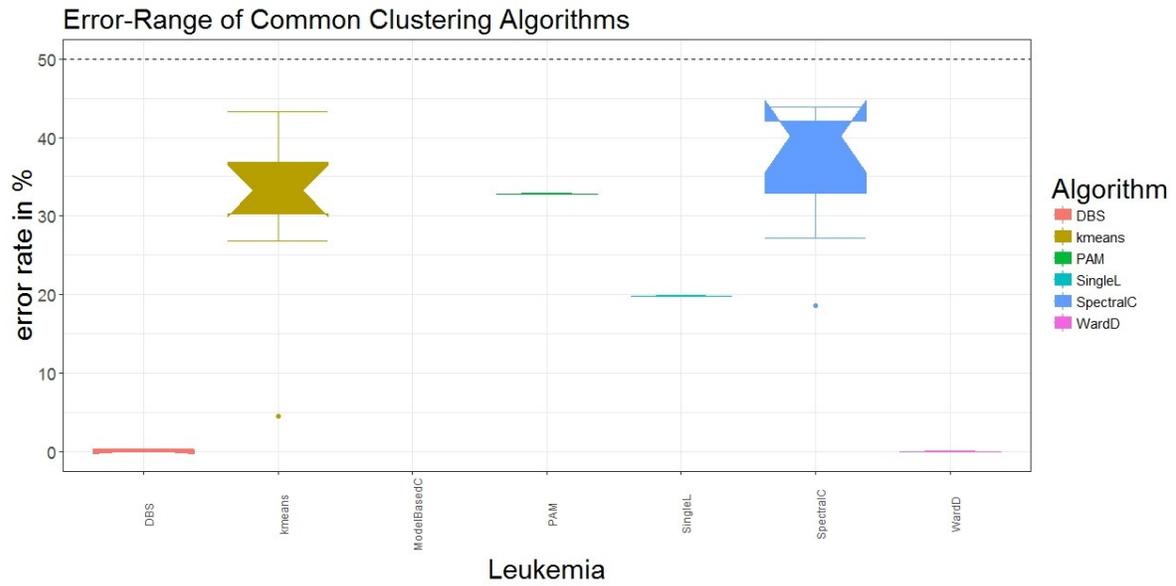

Figure A.1: The results from 10 trials of common clustering algorithms on Leukemia, shown as boxplots, where the level of error attributable to chance is 50% and the notch in each boxplot indicates the median.

Abbreviations: median (notch), single linkage (SL), Linde-Buzo-Gray algorithm (LBG-k-means), partitioning around medoids (PAM), mixture-of-Gaussians clustering (MoG), Databionic swarm (DBS).

## Supplementary Information B: Selected Applications of DBS on Real-World Data Sets Reproduce Cluster-Structures and Enable Knowledge Discovery

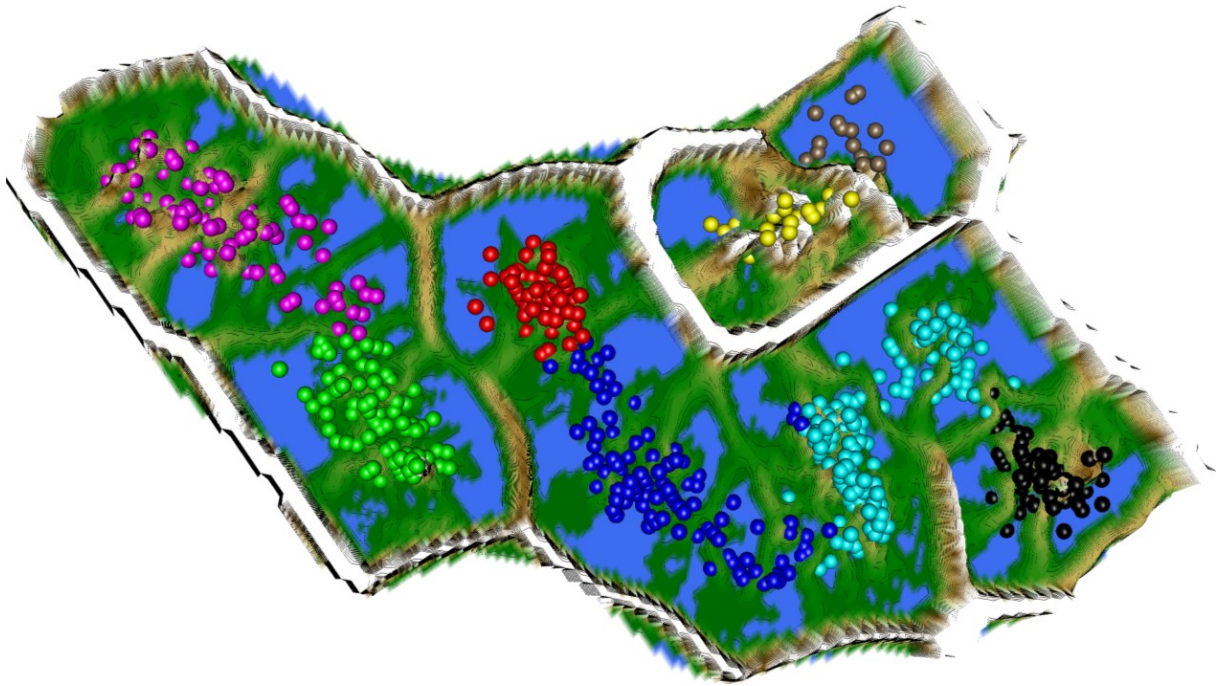

Figure B.2:    Topographic map of DBS clustering of 528 pain genes [Ultsch et al., 2016b]. The Cluster labeled in yellow consists of outliers [Thrun, 2018]. Clusters labeled in green and magenta, as well as the clusters labeled in blue and red,  are very similar to each other [Thrun, 2018].  The clusters are able to reproduce the knowledge about the functions of pain [Lötsch et al., 2013].

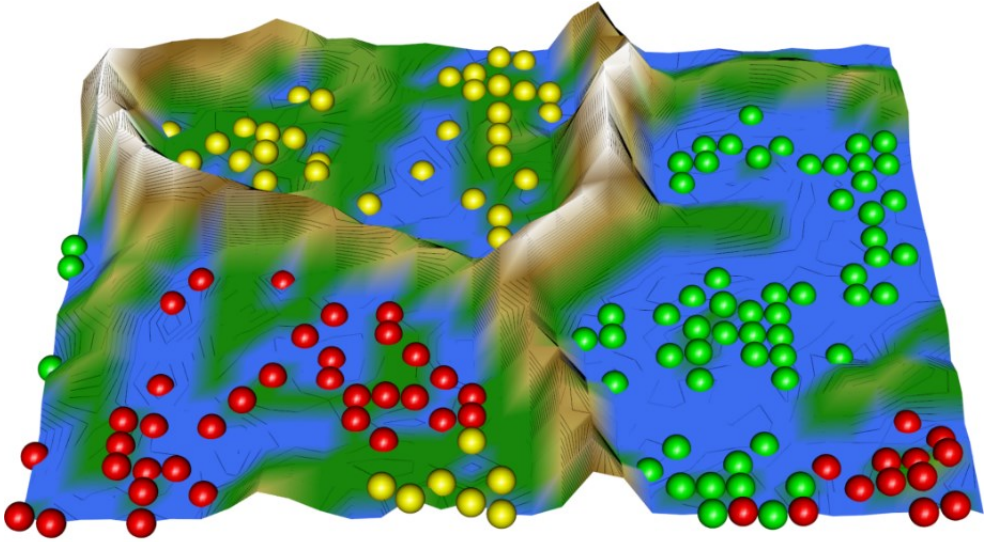

Figure B.3: Topographic map of high-dimensional data structure consisting of five single cold pain threshold measurements observed in 148 subjects [Weyer-Menkhoff et al., 2018]. Single measurements of the thresholds to tonic cold stimuli provided a 148 x 5 matrix[Weyer-Menkhoff et al., 2018]. The observation of data structures resembling the Gaussian mixture model-based prior classification was reproduced [Weyer-Menkhoff et al., 2018].

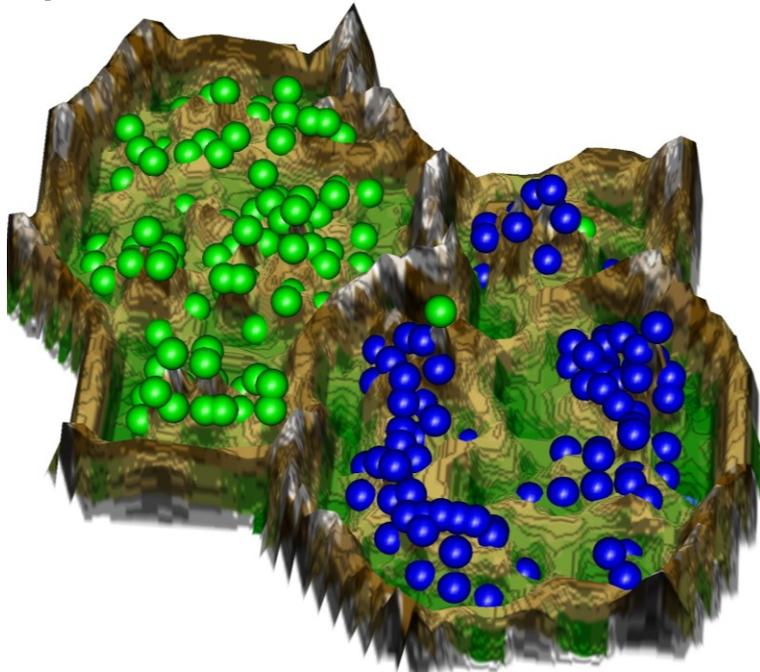

Figure B.4: Topographic map of the Swiss Banknotes data set [Flury/Riedwyl, 1988]. Two clusters are clearly visible, with two misplaced points. The clustering accuracy of the DBS is 99% [Thrun, 2018].

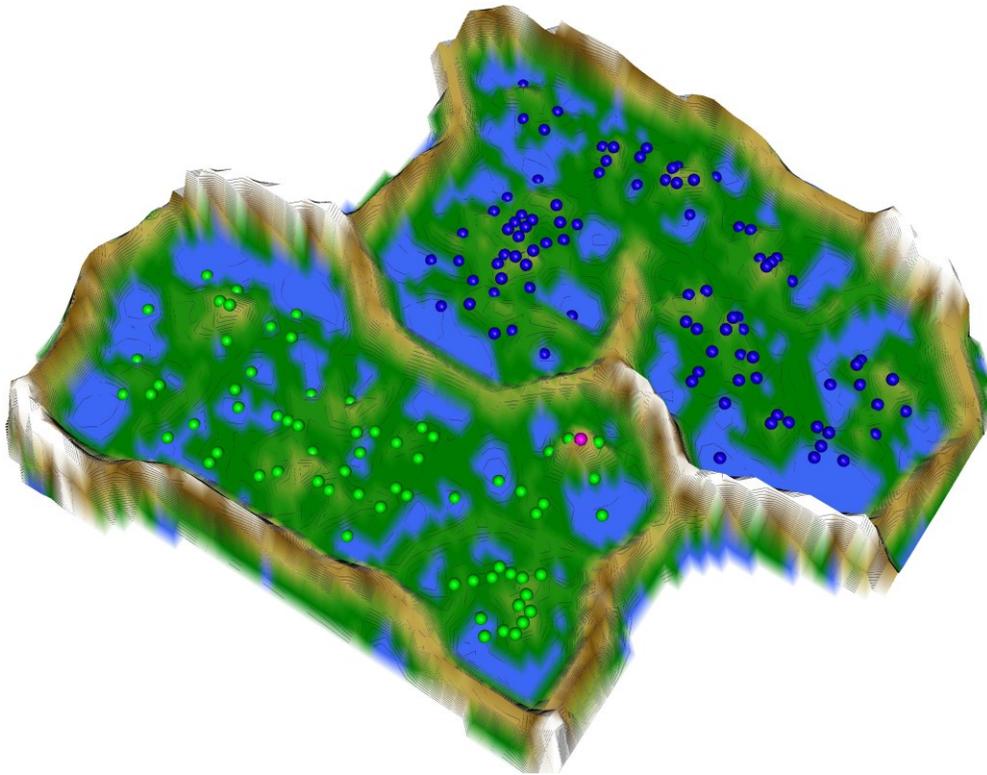

Figure B.5: Topographic map of the gross domestic products of 160 countries from 1970 to 2010 where the dynamic time warping distance is calculated between every two countries. The Topographic map shows two distinctive clusters. There is one outlier, colored in magenta. The rules deduced from CART show that the clusters are defined by an event occurring 2001 where the world economy was experiencing its first synchronized global recession in a quarter-century [Thrun, 2019]. Geographically, the first cluster consists of mostly African and Asian countries and a second cluster consists of mostly European and American countries [Thrun, 2019].

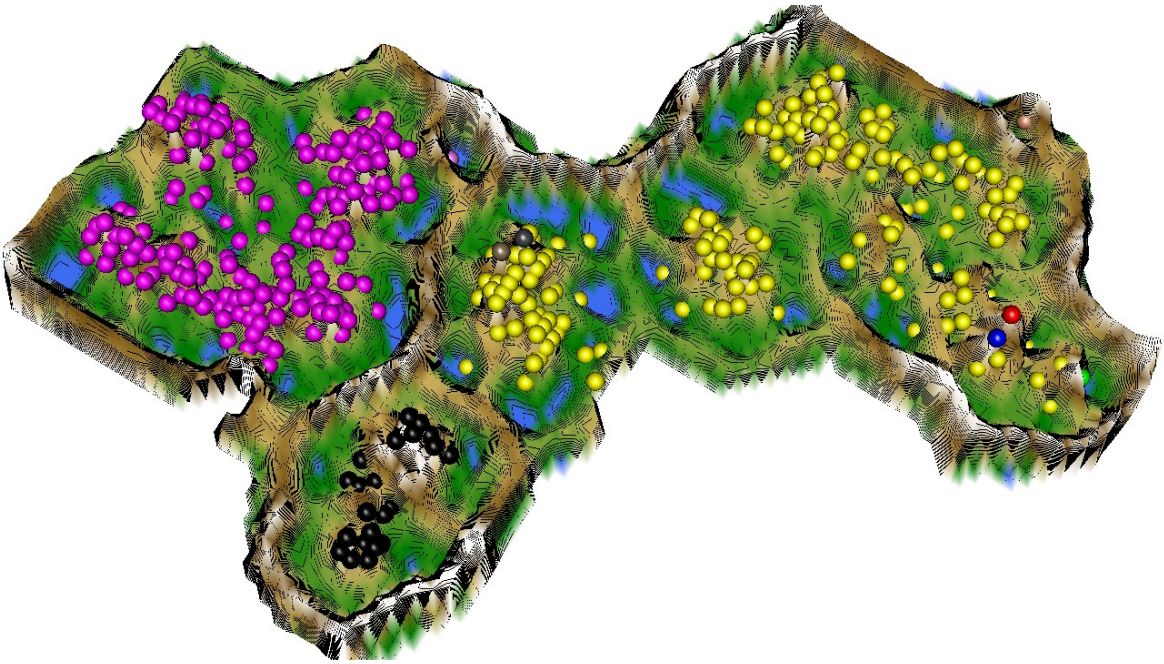

Figure B.6: The topographic map of the high-dimensional structures of the hydrology data set. The dataset contained in total 32,196 data points for 14 different variables in a 15 min frequency (15 min) of environmental measurements for a Catchment area (~3.7 km²) for two 2 years of measurements [Aubert et al., 2016]. For each day the measurements were aggregated by the sum[16] of all measurements for that day. The topographic map shows 3 clusters with distinctive types of days. The clusters enabled the domain expert to explain the biological and hydrological contributions leading to different states nitrate concentrations [Thrun et al., 2018].

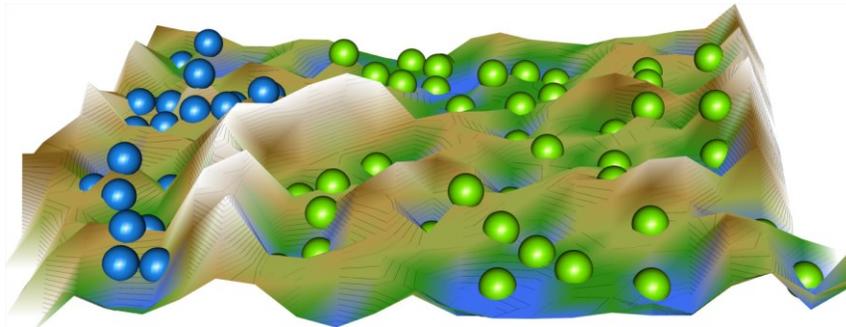

Figure B.7: Topographic map of high-dimensional data structures consisting of d = 31 gene loci analyzed in n = 75 subjects found in the TRPA1/TRPV1 NGS genotypes and its relation with the phenotypes[Kringel et al., 2018]. The cluster structure emerged from the separation of the data bots carrying the genetic information into two distinct groups, which was visually indicated by a "mountain range" on the topographic map. Cluster membership was unequally distributed among the phenotypes, i.e., the swarm-based cluster #1 comprising

---

[16] Recent research indicates that aggregation by mean instead of sum is preferable from the perspective of the domain expert resulting in substantial changes in the clustering and topographic map.

subjects carrying few variant alleles were clearly underrepresented in phenotype cluster (Gaussian) #1, i.e., among subjects with low heat hypersensitization response to capsaicin [Kringel et al., 2018].

## Supplementary Information C: Contingency Table for Tetragonula Compares DBS to Average Linkage

Table C.1: Contingency table of DBS clustering in rows versus H2014 ([Hennig 2014]) average linkage clustering in columns. Seven clusters can be reproduced. Total accuracy of DBS clustering in comparison to H2014 is 93%.

Abbreviations: $R\sum$ −Rowsum, R% - Rowpercentage, $C\sum$ −Columnsum, C% - Columnpercentage,

| H2014/ DBS | 1 | 2 | 3 | 4 | 5 | 6 | 7 | 8 | 9 | 10 | $R\sum$ | R% |
|---|---|---|---|---|---|---|---|---|---|---|---|---|
| 1 | 63 | 0 | 0 | 0 | 0 | 0 | 0 | 0 | 0 | 0 | 63 | 26,7 |
| 2 | 0 | 48 | 0 | 0 | 0 | 0 | 0 | 0 | 0 | 0 | 48 | 20,3 |
| 3 | 0 | 0 | 35 | 0 | 0 | 0 | 0 | 0 | 0 | 0 | 35 | 14,8 |
| 4 | 0 | 0 | 0 | 23 | 1 | 0 | 0 | 0 | 0 | 0 | 24 | 10,2 |
| 5 | 0 | 0 | 0 | 0 | 17 | 0 | 0 | 0 | 0 | 0 | 17 | 7,2 |
| 6 | 0 | 15 | 0 | 0 | 0 | 0 | 0 | 0 | 1 | 0 | 16 | 6,78 |
| 7 | 0 | 0 | 0 | 0 | 0 | 13 | 0 | 0 | 0 | 0 | 13 | 5,51 |
| 8 | 0 | 0 | 0 | 0 | 0 | 0 | 11 | 0 | 0 | 0 | 11 | 4,66 |
| 97 | 0 | 0 | 0 | 0 | 0 | 0 | 0 | 4 | 0 | 0 | 4 | 1,69 |
| 98 | 0 | 0 | 0 | 0 | 0 | 0 | 0 | 0 | 0 | 2 | 2 | 0,85 |
| 99 | 0 | 0 | 0 | 0 | 0 | 0 | 0 | 0 | 3 | 0 | 3 | 1,27 |
| $C\sum$ | 63 | 63 | 35 | 23 | 18 | 13 | 11 | 4 | 4 | 2 | 236 | 0 |
| C% | 26,7 | 26,7 | 14,8 | 9,75 | 7,63 | 5,51 | 4,66 | 1,69 | 1,69 | 0,85 | 0 | 100 |

# 9 References

[Abraham et al., 2006] **Abraham, A., Guo, H., & Liu, H.**: Swarm intelligence: foundations, perspectives and applications, In Nedjah, N. & Mourelle, L. d. M. (Eds.), *Swarm Intelligent Systems*, (pp. 3-25), Springer, **2006**.

[Aeberhard et al., 1992] **Aeberhard, S., Coomans, D., & De Vel, O.**: Comparison of classifiers in high dimensional settings, *Dept. Math. Statist., James Cook Univ., North Queensland, Australia, Tech. Rep, Vol.* (92-02), pp., **1992.**

[Anderson, 1935] **Anderson, E.**: The Irises of the Gaspé Peninsula, *Bulletin of the American Iris Society, Vol. 59*, pp. 2-5. **1935.**

[Aparna/Nair, 2014] **Aparna, K., & Nair, M. K.**: Enhancement of K-Means algorithm using ACO as an optimization technique on high dimensional data, Proc. Electronics and Communication Systems (ICECS), 2014 International Conference on, pp. 1-5, IEEE, **2014**.

[Arabie et al., 1996] **Arabie, P., Hubert, L. J., & De Soete, G.**: *Clustering and classification*, Singapore, World Scientific, ISBN: 9810212879, **1996**.

[Arumugam et al., 2005] **Arumugam, M. S., Chandramohan, A., & Rao, M.**: Competitive approaches to PSO algorithms via new acceleration co-efficient variant with mutation operators, Proc. Sixth International Conference on Computational Intelligence and Multimedia Applications (ICCIMA'05), pp. 225-230, IEEE, **2005**.

[Aubert et al., 2016] **Aubert, A. H., Thrun, M. C., Breuer, L., & Ultsch, A.**: Knowledge discovery from high-frequency stream nitrate concentrations: hydrology and biology contributions, *Scientific reports, Vol. 6*(31536), pp. doi 10.1038/srep31536, **2016.**

[Beckers et al., 1994] **Beckers, R., Holland, O., & Deneubourg, J.-L.**: From local actions to global tasks: Stigmergy and collective robotics, Proc. Artificial life IV, Vol. 181, pp. 189, **1994**.

[Bellman, 1957] **Bellman, R.**: Dynamic programming: Princeton Univ. press, Princeton, **1957**.

[Beni, 2004] **Beni, G.**: From swarm intelligence to swarm robotics, Proc. International Workshop on Swarm Robotics, pp. 1-9, Springer, **2004**.

[Beni/Wang, 1989] **Beni, G., & Wang, J.**: Swarm Intelligence in Cellular Robotic Systems, Proc. NATO Advanced Workshop on Robots and Biological Systems, Tuscany, Italy, **1989**.

[Benyus, 2002] **Benyus, J.**: Biomimicry: innovation inspired by design, New York: Harper Perennial, **2002**.

[Bogon, 2013] **Bogon, T.**:*Agentenbasierte Schwarmintelligenz,* (Phd Dissertation ), Springer-Verlag, Trier, Germany**, 2013**.

[Bonabeau et al., 1999] **Bonabeau, E., Dorigo, M., & Theraulaz, G.**: *Swarm intelligence: from natural to artificial systems*, New York, Oxford University Press, ISBN: 978-0-19-513159-8, **1999**.

[Bonabeau/Meyer, 2001] **Bonabeau, E., & Meyer, C.**: Swarm intelligence: A whole new way to think about business, *Harvard business review, Vol. 79*(5), pp. 106-115. **2001.**

[Bouveyron/Brunet-Saumard, 2014] **Bouveyron, C., & Brunet-Saumard, C.**: Model-based clustering of high-dimensional data: A review, *Computational Statistics & Data Analysis, Vol. 71*, pp. 52-78. **2014.**

[Brooks, 1991] **Brooks, R. A.**: Intelligence without representation, *Artificial intelligence, Vol. 47*(1), pp. 139-159. **1991.**

[Buhl et al., 2006] **Buhl, J., Sumpter, D. J., Couzin, I. D., Hale, J. J., Despland, E., Miller, E., & Simpson, S. J.**: From disorder to order in marching locusts, *Science, Vol. 312*(5778), pp. 1402-1406. **2006.**

[Chinchor, 1992] **Chinchor, N.**: MUC-4 evaluation metrics, Proc. Proceedings of the 4th conference on Message understanding, pp. 22-29, Association for Computational Linguistics, **1992**.

[Cormack, 1971] **Cormack, R. M.**: A review of classification, *Journal of the Royal Statistical Society. Series A (General), Vol.*, pp. 321-367. **1971.**

[Dasgupta/Gupta, 2003] **Dasgupta, S., & Gupta, A.**: An elementary proof of a theorem of Johnson and Lindenstrauss, *Random Structures & Algorithms, Vol. 22*(1), pp. 60-65. **2003.**

[Demartines/Hérault, 1995] **Demartines, P., & Hérault, J.**: CCA:" Curvilinear component analysis", Proc. 15° Colloque sur le traitement du signal et des images, Vol. 199, GRETSI, Groupe d'Etudes du Traitement du Signal et des Images, France 18-21 September, **1995**.

[Deneubourg et al., 1991] **Deneubourg, J.-L., Goss, S., Franks, N., Sendova-Franks, A., Detrain, C., & Chrétien, L.**: The dynamics of collective sorting robot-like ants and ant-like robots, Proc. Proceedings of the first international conference on simulation of adaptive behavior on From animals to animats, pp. 356-363, **1991**.

[Dijkstra, 1959] **Dijkstra, E. W.**: A note on two problems in connexion with graphs, *Numerische mathematik, Vol. 1*(1), pp. 269-271. **1959.**

[Duda et al., 2001] **Duda, R. O., Hart, P. E., & Stork, D. G.**: *Pattern Classification*, (Second Edition ed.), Ney York, USA, John Wiley & Sons, ISBN: 0-471-05669-3, **2001**.

[Eberhart et al., 2001] **Eberhart, R. C., Shi, Y., & Kennedy, J.**: Swarm Intelligence (The Morgan Kaufmann Series in Evolutionary Computation)*, Vol.*, pp., **2001.**

[Erwin et al., 1992] **Erwin, E., Obermayer, K., & Schulten, K.**: Self-organizing maps: Stationary states, metastability and convergence rate, *Biological cybernetics, Vol. 67*(1), pp. 35-45. **1992.**

[Esmin et al., 2015] **Esmin, A. A., Coelho, R. A., & Matwin, S.**: A review on particle swarm optimization algorithm and its variants to clustering high-dimensional data, *Artificial Intelligence Review, Vol. 44*(1), pp. 23-45. **2015.**

[Ester et al., 1996] **Ester, M., Kriegel, H.-P., Sander, J., & Xu, X.**: A Density-Based Algorithm for Discovering Clusters in Large Spatial Databases with Noise, Proc. Second International Conference on Knowledge Discovery and Data Mining (KDD 96), Vol. 96, pp. 226-231, AAAI Press, Portland, Oregon August, **1996**.

[Everitt et al., 2001] **Everitt, B. S., Landau, S., & Leese, M.**: *Cluster analysis*, (McAllister, L. Ed. Fourth Edition ed.), London, Arnold, ISBN: 978-0-340-76119-9, **2001**.

[Fathian/Amiri, 2008] **Fathian, M., & Amiri, B.**: A honeybee-mating approach for cluster analysis, *The International Journal of Advanced Manufacturing Technology, Vol. 38*(7-8), pp. 809-821. **2008.**

[Feynman et al., 2007] **Feynman, R. P., Leighton, R. B., & Sands, M.**: *Mechanik, Strahlung, Wärme*, (Köhler, K.-H., Schröder, K.-E. & Beckmann, W. B. P., Trans. 5th Edition: Definitive Edition ed. Vol. 1), Munich, Germany, Oldenbourg Verlag, ISBN: 978-3-486-58108-9, **2007**.

[Florek et al., 1951]  **Florek, K., Łukaszewicz, J., Perkal, J., Steinhaus, H., & Zubrzycki, S.**: Sur la liaison et la division des points d'un ensemble fini, Proc. Colloquium Mathematicae, Vol. 2, pp. 282-285, Institute of Mathematics Polish Academy of Sciences, **1951**.

[Flury/Riedwyl, 1988]  **Flury, B., & Riedwyl, H.**: *Multivariate statistics, a practical approach*, London, Chapman and Hall, ISBN, **1988**.

[Forest, 1990]  **Forest, S.**: Emergent computation: self-organizing, collective, and cooperative phenomena in natural and artificial computing networks, *Physica D, Vol. 42*, pp. 1-11. **1990.**

[Fraley/Raftery, 2002]  **Fraley, C., & Raftery, A. E.**: Model-based clustering, discriminant analysis, and density estimation, *Journal of the American Statistical Association, Vol. 97*(458), pp. 611-631. **2002.**

[Fraley/Raftery, 2006]  **Fraley, C., & Raftery, A. E.**MCLUST version 3: an R package for normal mixture modeling and model-based clustering, Technical Report No. 504,Department of Statistics, University of Washington, Report No. Seattle, USA, **2006**.

[Franck et al., 2004]  **Franck, P., Cameron, E., Good, G., RASPLUS, J. Y., & Oldroyd, B.**: Nest architecture and genetic differentiation in a species complex of Australian stingless bees, *Molecular Ecology, Vol. 13*(8), pp. 2317-2331. **2004**.

[Garnier et al., 2007]  **Garnier, S., Gautrais, J., & Theraulaz, G.**: The biological principles of swarm intelligence, *Swarm Intelligence, Vol. 1*(1), pp. 3-31. **2007**.

[Geman et al., 1992]  **Geman, S., Bienenstock, E., & Doursat, R.**: Neural networks and the bias/variance dilemma, *Neural Computation, Vol. 4*(1), pp. 1-58. **1992.**

[Gigerenzer/Brighton, 2009]  **Gigerenzer, G., & Brighton, H.**: Homo heuristicus: Why biased minds make better inferences, *Topics in cognitive science, Vol. 1*(1), pp. 107-143. **2009.**

[Giraldo et al., 2011]  **Giraldo, L. F., Lozano, F., & Quijano, N.**: Foraging theory for dimensionality reduction of clustered data, *Machine Learning, Vol. 82*(1), pp. 71-90. **2011.**

[Goldstein, 1999]  **Goldstein, J.**: Emergence as a construct: History and issues, *Emergence, Vol. 1*(1), pp. 49-72. **1999.**

[Grassé, 1959]  **Grassé, P.-P.**: La reconstruction du nid et les coordinations interindividuelles chezBellicositermes natalensis etCubitermes sp. la théorie de la stigmergie: Essai d'interprétation du comportement des termites constructeurs, *Insectes sociaux, Vol. 6*(1), pp. 41-80. **1959**.

[Grosan et al., 2006]  **Grosan, C., Abraham, A., & Chis, M.**: Swarm intelligence in data mining, *Swarm Intelligence in Data Mining*, (pp. 1-20), Springer, **2006**.

[Handl et al., 2003]  **Handl, J., Knowles, J., & Dorigo, M.**: Ant-based clustering: a comparative study of its relative performance with respect to k-means, average link and 1d-som, *Design and application of hybrid intelligent systems, Vol.*, pp. 204-213. **2003.**

[Handl et al., 2006]  **Handl, J., Knowles, J., & Dorigo, M.**: Ant-based clustering and topographic mapping, *Artificial Life, Vol. 12*(1), pp. 35-62. **2006.**

[Handl et al., 2005]  **Handl, J., Knowles, J., & Kell, D. B.**: Computational cluster validation in post-genomic data analysis, *Bioinformatics, Vol. 21*(15), pp. 3201-3212. **2005.**

[Hartigan, 1981] **Hartigan, J. A.**: Consistency of single linkage for high-density clusters, *Journal of the American Statistical Association, Vol. 76*(374), pp. 388-394. **1981.**

[Haug/Koch, 2004] **Haug, H., & Koch, S. W.**: *Quantum theory of the optical and electronic properties of semiconductors*, (Edition, F. Ed. Vol. 5), Singapore, World Scientific, ISBN: 13 978-981-283-883-4, **2004**.

[Havens et al., 2008] **Havens, T. C., Spain, C. J., Salmon, N. G., & Keller, J. M.**: Roach infestation optimization, Proc. Swarm Intelligence Symposium, 2008. SIS 2008. IEEE, pp. 1-7, IEEE, **2008**.

[Hennig, 2014] **Hennig, C.**: How many bee species? A case study in determining the number of clusters, *Data Analysis, Machine Learning and Knowledge Discovery*, (pp. 41-49), Springer, **2014**.

[Hennig, 2015] **Hennig, C., et al. (Hg.)**: *Handbook of cluster analysis*, New York, USA, Chapman&Hall/CRC Press, ISBN: 9781466551893, **2015**.

[Herrmann, 2011] **Herrmann, L.**:*Swarm-Organized Topographic Mapping,* (Doctoral dissertation), Philipps-Universität Marburg, Marburg**, 2011**.

[Herrmann/Ultsch, 2008a] **Herrmann, L., & Ultsch, A.**: The architecture of ant-based clustering to improve topographic mapping, *Ant Colony Optimization and Swarm Intelligence*, (pp. 379-386), Springer, **2008a**.

[Herrmann/Ultsch, 2008b] **Herrmann, L., & Ultsch, A.**: Explaining Ant-Based Clustering on the basis of Self-Organizing Maps, Proc. ESANN, pp. 215-220, Citeseer, **2008b**.

[Herrmann/Ultsch, 2009] **Herrmann, L., & Ultsch, A.**: Clustering with swarm algorithms compared to emergent SOM, Proc. International Workshop on Self-Organizing Maps, pp. 80-88, Springer, **2009**.

[Hinton/Roweis, 2002] **Hinton, G. E., & Roweis, S. T.**: Stochastic neighbor embedding, Proc. Advances in neural information processing systems, pp. 833-840, **2002**.

[Hunklinger, 2009] **Hunklinger, S.**: *Festkörperphysik*, Oldenbourg Verlag, ISBN: 3486596411, **2009**.

[Jafar/Sivakumar, 2010] **Jafar, O. M., & Sivakumar, R.**: Ant-based clustering algorithms: A brief survey, *International Journal of Computer Theory and Engineering, Vol. 2*(5), pp. 787. **2010.**

[Jain/Dubes, 1988] **Jain, A. K., & Dubes, R. C.**: *Algorithms for Clustering Data*, Englewood Cliffs, New Jersey, USA, Prentice Hall College Div, ISBN: 9780130222787, **1988**.

[Janich/Duncker, 2011] **Janich, P., & Duncker, H.-R.**: *Emergenz-Lückenbüssergottheit für Natur- und Geisteswissenschaften*, F. Steiner, ISBN: 978-3-515-09871-7, **2011**.

[Jennings et al., 1998] **Jennings, N. R., Sycara, K., & Wooldridge, M.**: A roadmap of agent research and development, *Autonomous agents and multi-agent systems, Vol. 1*(1), pp. 7-38. **1998.**

[Kämpf/Ultsch, 2006] **Kämpf, D., & Ultsch, A.**: An Overview of Artificial Life Approaches for Clustering, *From Data and Information Analysis to Knowledge Engineering*, (pp. 486-493), Springer, **2006**.

[Karaboga, 2005] **Karaboga, D.**An idea based on honey bee swarm for numerical optimization,Technical report-tr06, Erciyes university, engineering faculty, computer engineering department, Report No., **2005**.

[Karaboga/Akay, 2009]  **Karaboga, D., & Akay, B.**: A survey: algorithms simulating bee swarm intelligence, *Artificial Intelligence Review, Vol. 31*(1-4), pp. 61-85. **2009.**

[Karaboga/Ozturk, 2011]  **Karaboga, D., & Ozturk, C.**: A novel clustering approach: Artificial Bee Colony (ABC) algorithm, *Applied Soft Computing, Vol. 11*(1), pp. 652-657. **2011.**

[Kaufman/Rousseeuw, 1990]  **Kaufman, L., & Rousseeuw, P. J.**: Partitioning around medoids (program pam), *Finding groups in data: an introduction to cluster analysis, Vol.*, pp. 68-125. **1990.**

[Kaufman/Rousseeuw, 2005]  **Kaufman, L., & Rousseeuw, P. J.**: *Finding groups in data: An introduction to cluster analysis*, Hoboken New York, John Wiley & Sons Inc, ISBN: 0-471-73578-7, **2005**.

[Kaur/Rohil, 2015]  **Kaur, P., & Rohil, H.**: Applications of Swarm Intelligence in Data Clustering: A Comprehensive Review, *International Journal of Research in Advent Technology (IJRAT), Vol. 3*(4), pp. 85-95. **2015.**

[Kelso, 1997]  **Kelso, J. A. S.**: *Dynamic patterns: The self-organization of brain and behavior*, Cambridge, Massachusetts, London, England, MIT press, ISBN: 0262611317, **1997**.

[Kennedy/Eberhart, 1995]  **Kennedy, J., & Eberhart, R.**: *Particle Swarm Optimization*, IEEE International Conference on Neural Networks, Vol. 4, pp. 1942-1948, IEEE Service Center,, Piscataway, **1995.**

[Kleinberg, 2003]  **Kleinberg, J.**: An impossibility theorem for clustering, Proc. Advances in neural information processing systems, Vol. 15, pp. 463-470, MIT Press, Vancouver, British Columbia, Canada December 9-14, **2003**.

[Kohonen, 1982]  **Kohonen, T.**: Self-organized formation of topologically correct feature maps, *Biological cybernetics, Vol. 43*(1), pp. 59-69. **1982.**

[Kringel et al., 2018]  **Kringel, D., Geisslinger, G., Resch, E., Oertel, B. G., Thrun, M. C., Heinemann, S., & Lötsch, J.**: Machine-learned analysis of the association of next-generation sequencing based human TRPV1 and TRPA1 genotypes with the sensitivity to heat stimuli and topically applied capsaicin, *Pain, Vol. 159*(7), pp. 1366–1381. doi 10.1097/j.pain.0000000000001222, **2018.**

[Legg/Hutter, 2007]  **Legg, S., & Hutter, M.**: A collection of definitions of intelligence, *Frontiers in Artificial Intelligence and applications, Vol. 157*, pp. 17. **2007.**

[Li/Xiao, 2008]  **Li, J., & Xiao, X.**: Multi-swarm and multi-best particle swarm optimization algorithm, Proc. Intelligent Control and Automation, 2008. WCICA 2008. 7th World Congress on, pp. 6281-6286, IEEE, **2008**.

[Linde et al., 1980]  **Linde, Y., Buzo, A., & Gray, R.**: An algorithm for vector quantizer design, *IEEE Transactions on communications, Vol. 28*(1), pp. 84-95. **1980.**

[Lötsch et al., 2013]  **Lötsch, J., Doehring, A., Mogil, J. S., Arndt, T., Geisslinger, G., & Ultsch, A.**: Functional genomics of pain in analgesic drug development and therapy, *Pharmacology & therapeutics, Vol. 139*(1), pp. 60-70. **2013.**

[Lötsch/Ultsch, 2014]  **Lötsch, J., & Ultsch, A.**: Exploiting the Structures of the U-Matrix, in Villmann, T., Schleif, F.-M., Kaden, M. & Lange, M. (eds.), Proc. Advances in Self-Organizing Maps and Learning Vector Quantization, pp. 249-257, Springer International Publishing, Mittweida, Germany July 2–4, **2014**.

[Lumer/Faieta, 1994] **Lumer, E. D., & Faieta, B.**: Diversity and adaptation in populations of clustering ants, Proc. Proceedings of the third international conference on Simulation of adaptive behavior: from animals to animats 3: from animals to animats 3, pp. 501-508, MIT Press, **1994**.

[MacQueen, 1967] **MacQueen, J.**: Some methods for classification and analysis of multivariate observations, Proc. Proceedings of the fifth Berkeley symposium on mathematical statistics and probability, Vol. 1, pp. 281-297, Oakland, CA, USA., **1967**.

[Marinakis et al., 2007] **Marinakis, Y., Marinaki, M., & Matsatsinis, N.**: A hybrid clustering algorithm based on honey bees mating optimization and greedy randomized adaptive search procedure, Proc. International Conference on Learning and Intelligent Optimization, pp. 138-152, Springer, **2007**.

[Martens et al., 2011] **Martens, D., Baesens, B., & Fawcett, T.**: Editorial survey: swarm intelligence for data mining, *Machine Learning, Vol. 82*(1), pp. 1-42. **2011.**

[McDonnell, 1995] **McDonnell, R.**: *International GIS dictionary* Retrieved from http://support.esri.com/other-resources/gis-dictionary/term/boundary%20effect, 30.11.2016 11:10, **1995**.

[Menéndez et al., 2014] **Menéndez, H. D., Otero, F. E., & Camacho, D.**: MACOC: a medoid-based ACO clustering algorithm, Proc. International Conference on Swarm Intelligence, pp. 122-133, Springer, **2014**.

[Mirkin, 2005] **Mirkin, B. G.**: *Clustering: a data recovery approach*, Boca Raton, FL, USA, Chapman&Hall/CRC, ISBN: 978-1-58488-534-4, **2005**.

[Mlot et al., 2011] **Mlot, N. J., Tovey, C. A., & Hu, D. L.**: Fire ants self-assemble into waterproof rafts to survive floods, *Proceedings of the National Academy of Sciences, Vol. 108*(19), pp. 7669-7673. **2011.**

[Nash, 1950] **Nash, J. F.**: Equilibrium points in n-person games, *Proc. Nat. Acad. Sci. USA, Vol. 36*(1), pp. 48-49. **1950.**

[Nash, 1951] **Nash, J. F.**: Non-cooperative games, *Annals of mathematics, Vol.*, pp. 286-295. **1951.**

[Neumann/Morgenstern, 1953] **Neumann, L. J., & Morgenstern, O.**: *Theory Of Games And Economic Behavior* (Third Edition ed. Vol. 60), Princeton, USA, Princeton University Press, ISBN, **1953**.

[Ng et al., 2002] **Ng, A. Y., Jordan, M. I., & Weiss, Y.**: On spectral clustering: Analysis and an algorithm, *Advances in neural information processing systems, Vol. 2*, pp. 849-856. **2002.**

[Nisan et al., 2007] **Nisan, N., Roughgarden, T., Tardos, E., & Vazirani, V. V.**: *Algorithmic Game Theory*, (Nisan, N. Ed.), New York, USA, Cambridge University Press, ISBN: 978-0-521-87282-9, **2007**.

[Nybo et al., 2007] **Nybo, K., Venna, J., & Kaski, S.**: The self-organizing map as a visual information retrieval method, *Vol.*, pp., **2007.**

[Omar et al., 2013] **Omar, E., Badr, A., & Hegazy, A. E.-F.**: Hybrid AntBased Clustering Algorithm with Cluster Analysis Techniques, Proc. Journal of Computer Science, Vol. 9, pp. 780-793, Citeseer, **2013**.

[Ouadfel/Batouche, 2007] **Ouadfel, S., & Batouche, M.**: An Efficient Ant Algorithm for Swarm-Based Image Clustering 1, *Journal of Computer Science, Vol. 3*(3), pp. 2-167, 162. **2007.**

[Parpinelli/Lopes, 2011]  **Parpinelli, R. S., & Lopes, H. S.**: New inspirations in swarm intelligence: a survey, *International Journal of Bio-Inspired Computation, Vol. 3*(1), pp. 1-16. **2011.**

[Pasquier, 1987]  **Pasquier, V.**: Lattice derivation of modular invariant partition functions on the torus, *Journal of Physics A: Mathematical and General, Vol. 20*(18), pp. L1229. **1987.**

[Passino, 2013]   **Passino, K. M.**: Modeling and Cohesiveness Analysis of Midge Swarms, *International Journal of Swarm Intelligence Research (IJSIR), Vol. 4*(4), pp. 1-22. **2013.**

[Pham et al., 2007]  **Pham, D., Otri, S., Afify, A., Mahmuddin, M., & Al-Jabbouli, H.**: Data clustering using the bees algorithm, Proc. Proceedings of 40th CIRP international manufacturing systems seminar, **2007**.

[Rana et al., 2011]  **Rana, S., Jasola, S., & Kumar, R.**: A review on particle swarm optimization algorithms and their applications to data clustering, *Artificial Intelligence Review, Vol. 35*(3), pp. 211-222. **2011.**

[Reynolds, 1987]  **Reynolds, C. W.**: Flocks, herds and schools: A distributed behavioral model, *ACM SIGGRAPH computer graphics, Vol. 21*(4), pp. 25-34. **1987.**

[Şahin, 2004]  **Şahin, E.**: Swarm robotics: From sources of inspiration to domains of application, Proc. International workshop on swarm robotics, pp. 10-20, Springer, **2004**.

[Schelling, 1969]  **Schelling, T. C.**: Models of segregation, *The American Economic Review, Vol. 59*(2), pp. 488-493. **1969.**

[Schneirla, 1971]  **Schneirla, T.**: *Army ants, a study in social organization*, San Francisco, USA, W.H. Freeman and Company, ISBN: 0-7167-0933-3, **1971**.

[Shelokar et al., 2004]  **Shelokar, P., Jayaraman, V. K., & Kulkarni, B. D.**: An ant colony approach for clustering, *Analytica Chimica Acta, Vol. 509*(2), pp. 187-195. **2004.**

[Stephens/Krebs, 1986]  **Stephens, D. W., & Krebs, J. R.**: *Foraging theory*, New Jersey, USA, Princeton University Press, ISBN: 0691084424, **1986**.

[Tan et al., 2006]  **Tan, S. C., Ting, K. M., & Teng, S. W.**: Reproducing the results of ant-based clustering without using ants, Proc. 2006 IEEE International Conference on Evolutionary Computation, pp. 1760-1767, IEEE, **2006**.

[Theodoridis/Koutroumbas, 2009]  **Theodoridis, S., & Koutroumbas, K.**: *Pattern Recognition*, (Fourth Edition ed.), Canada, Elsevier, ISBN: 978-1-59749-272-0, **2009**.

[Thrun, 2017]  **Thrun, M. C.:** DatabionicSwarm (Version 1.0), Marburg. R package, requires CRAN packages: Rcpp, RcppArmadillo, deldir, Suggests: plotrix, geometry, sp, spdep, AdaptGauss, ABCanalysis, parallel, Retrieved from [https://cran.r-project.org/web/packages/DataBionicSwarm/index.html](https://cran.r-project.org/web/packages/DataBionicSwarm/index.html), **2017.**

[Thrun, 2018]  **Thrun, M. C.**: *Projection Based Clustering through Self-Organization and Swarm Intelligence*, (Ultsch, A. & Hüllermeier, E. Eds., 10.1007/978-3-658-20540-9), Extended doctoral dissertation, Heidelberg, Springer, ISBN: 978-3658205393, **2018**.

[Thrun, 2019]   **Thrun, M. C.**: Cluster Analysis of Per Capita Gross Domestic Products, *Entrepreneurial Business and Economics Review (EBER), Vol. 7*(1), pp. 217-231. doi 10.15678/EBER.2019.070113, **2019.**

[Thrun et al., 2018]   **Thrun, M. C., Breuer, L., & Ultsch, A.**: Knowledge discovery from low-frequency stream nitrate concentrations: hydrology and biology contributions, Proc. European Conference on Data Analysis (ECDA), pp. 46-47, Paderborn, Germany, **2018**.

[Thrun et al., 2020]   **Thrun, M. C., Gehlert, T., & Ultsch, A.**: Analyzing the Fine Structure of Distributions, *PloS one, Vol. 15*(10), pp. e0238835. doi 10.1371/journal.pone.0238835 **2020.**

[Thrun et al., 2016]   **Thrun, M. C., Lerch, F., Lötsch, J., & Ultsch, A.**: *Visualization and 3D Printing of Multivariate Data of Biomarkers*, in Skala, V. (Ed.), International Conference in Central Europe on Computer Graphics, Visualization and Computer Vision (WSCG), Vol. 24, pp. 7-16, Plzen, http://wscg.zcu.cz/wscg2016/short/A43-full.pdf, **2016.**

[Thrun/Ultsch, 2020a]   **Thrun, M. C., & Ultsch, A.**: Clustering Benchmark Datasets Exploiting the Fundamental Clustering Problems, *Data in Brief, Vol. 30*(C), pp. 105501. doi 10.1016/j.dib.2020.105501, **2020a.**

[Thrun/Ultsch, 2020b]   **Thrun, M. C., & Ultsch, A.**: Uncovering High-Dimensional Structures of Projections from Dimensionality Reduction Methods, *MethodsX, Vol. 7*, pp. 101093. doi 10.1016/j.mex.2020.101093, **2020b.**

[Timm, 2006]   **Timm, I. J.**: Strategic management of autonomous software systems, *TZI-Bericht Center for Computing Technologies, University of Bremen, Bremen, Vol.*, pp., **2006.**

[Toussaint, 1980]   **Toussaint, G. T.**: The relative neighbourhood graph of a finite planar set, *Pattern Recognition, Vol. 12*(4), pp. 261-268. **1980.**

[Tsai et al., 2004]   **Tsai, C.-F., Tsai, C.-W., Wu, H.-C., & Yang, T.**: ACODF: a novel data clustering approach for data mining in large databases, *Journal of Systems and Software, Vol. 73*(1), pp. 133-145. **2004.**

[Uber_Pix, 2015]   **Uber_Pix.** A spherical school of fish, In 567289_1280_1024.jpg (Ed.), (Vol. 1280x1024), https://twitter.com/Uber_Pix/status/614068525766995969, twitter, **2015**.

[Ultsch, 1999]   **Ultsch, A.**: Data mining and knowledge discovery with emergent self-organizing feature maps for multivariate time series, In Oja, E. & Kaski, S. (Eds.), *Kohonen maps*, (1 ed., pp. 33-46), Elsevier, **1999**.

[Ultsch, 2000a]   **Ultsch, A.**: *Clustering with DataBots*, Int. Conf. Advances in Intelligent Systems Theory and Applications (AISTA), pp. p. 99-104, IEEE ACT Section, Canberra, Australia, **2000a.**

[Ultsch, 2000b]   **Ultsch, A.**: Visualisation and Classification with Artificial Life, *Data Analysis, Classification, and Related Methods*, (pp. 229-234), Springer, **2000b.**

[Ultsch, 2007]   **Ultsch, A.**: Emergence in Self-Organizing Feature Maps, Proc. 6th Workshop on Self-Organizing Maps (WSOM 07), 10.2390/biecoll-wsom2007-114, pp. 1-7, University Library of Bielefeld, Bielefeld, Germany, **2007**.

[Ultsch et al., 2016a]   **Ultsch, A., Behnisch, M., & Lötsch, J.**: ESOM Visualizations for Quality Assessment in Clustering, In Merényi, E., Mendenhall, J. M. & O'Driscoll, P. (Eds.), *Advances in Self-Organizing Maps and Learning Vector Quantization: Proceedings of the 11th International Workshop WSOM 2016, Houston, Texas, USA, January 6-8, 2016*, (10.1007/978-3-319-28518-4_3pp. 39-48), Cham, Springer International Publishing, **2016a.**

[Ultsch/Herrmann, 2005] **Ultsch, A., & Herrmann, L.**: The architecture of emergent self-organizing maps to reduce projection errors, in Verleysen, M. (Ed.), Proc. European Symposium on Artificial Neural Networks (ESANN), pp. 1-6, Bruges, Belgium April, **2005**.

[Ultsch et al., 2016b] **Ultsch, A., Kringel, D., Kalso, E., Mogil, J. S., & Lötsch, J.**: A data science approach to candidate gene selection of pain regarded as a process of learning and neural plasticity, *Pain, Vol.*, pp., **2016b.**

[Ultsch/Lötsch, 2017] **Ultsch, A., & Lötsch, J.**: Machine-learned cluster identification in high-dimensional data, *Journal of biomedical informatics, Vol. 66*(C), pp. 95-104. **2017.**

[Ultsch/Thrun, 2017] **Ultsch, A., & Thrun, M. C.**: *Credible Visualizations for Planar Projections*, in Cottrell, M. (Ed.), 12th International Workshop on Self-Organizing Maps and Learning Vector Quantization, Clustering and Data Visualization (WSOM)*, 10.1109/WSOM.2017.8020010, pp. 1-5, IEEE, Nany, France, **2017.**

[Van der Merwe/Engelbrecht, 2003] **Van der Merwe, D., & Engelbrecht, A. P.**: Data clustering using particle swarm optimization, Proc. Evolutionary Computation, 2003. CEC'03. The 2003 Congress on, Vol. 1, pp. 215-220, IEEE, **2003**.

[Van Rijsbergen, 1979] **Van Rijsbergen, C.**: Information retrieval/CJ van Rijsbergen, Butterworths, London, **1979**.

[Venna/Kaski, 2007] **Venna, J., & Kaski, S.**: Comparison of visualization methods for an atlas of gene expression data sets, *Information Visualization, Vol. 6*(2), pp. 139-154. **2007.**

[Venna et al., 2010] **Venna, J., Peltonen, J., Nybo, K., Aidos, H., & Kaski, S.**: Information retrieval perspective to nonlinear dimensionality reduction for data visualization, *The Journal of Machine Learning Research, Vol. 11*, pp. 451-490. **2010.**

[Wang et al., 2007] **Wang, Z., Sun, X., & Zhang, D.**: A PSO-based classification rule mining algorithm, Proc. International Conference on Intelligent Computing, pp. 377-384, Springer, **2007**.

[Ward Jr, 1963] **Ward Jr, J. H.**: Hierarchical grouping to optimize an objective function, *Journal of the American Statistical Association, Vol. 58*(301), pp. 236-244. **1963.**

[Weyer-Menkhoff et al., 2018] **Weyer-Menkhoff, I., Thrun, M. C., & Lötsch, J.**: Machine-learned analysis of quantitative sensory testing responses to noxious cold stimulation in healthy subjects, *European Journal of Pain, Vol. 22*(5), pp. 862-874. doi 10.1002/ejp.1173, **2018.**

[Wolpert, 1996] **Wolpert, D. H.**: The lack of a priori distinctions between learning algorithms, *Neural Computation, Vol. 8*(7), pp. 1341-1390. **1996.**

[Wong et al., 2014] **Wong, K.-C., Peng, C., Li, Y., & Chan, T.-M.**: Herd clustering: A synergistic data clustering approach using collective intelligence, *Applied Soft Computing, Vol. 23*, pp. 61-75. **2014.**

[Yang, 2009] **Yang, X.-S.**: Firefly algorithms for multimodal optimization, Proc. International Symposium on Stochastic Algorithms, pp. 169-178, Springer, **2009**.

[Yang/He, 2013] **Yang, X.-S., & He, X.**: Bat algorithm: literature review and applications, *International Journal of Bio-Inspired Computation, Vol. 5*(3), pp. 141-149. **2013.**

[Zhong, 2010] **Zhong, Y.**: Advanced intelligence: definition, approach, and progresses, *International Journal of Advanced Intelligence, Vol. 2*(1), pp. 15-23. **2010.**

[Zou et al., 2010]  **Zou, W., Zhu, Y., Chen, H., & Sui, X.**: A clustering approach using cooperative artificial bee colony algorithm, *Discrete Dynamics in Nature and Society, Vol. 2010*, pp., **2010.**